\documentclass[10pt,twocolumn,letterpaper]{article}

\usepackage{iccv}
\usepackage{times}
\usepackage{epsfig}
\usepackage{graphicx}
\usepackage{amsmath}
\usepackage{amssymb}
\usepackage{booktabs}
\usepackage{array}
\usepackage{multirow}
\usepackage{color}
\usepackage{colortbl}
\usepackage{framed}
\usepackage{bm}
\usepackage{bbm}
\usepackage{xspace}
\usepackage{enumitem}
\usepackage{cite}
\usepackage{xparse}
\usepackage{xcolor}
\usepackage{algorithm}
\usepackage{lipsum}
\usepackage{listings}
\usepackage{url}
\usepackage{pifont}
\usepackage{subcaption}
\usepackage[accsupp]{axessibility}

\usepackage[pagebackref=true,breaklinks=true,letterpaper=true,colorlinks,bookmarks=false]{hyperref}

\definecolor{Gray}{gray}{0.9}
\definecolor{LightCyan}{rgb}{0.88,0.95,1}
\definecolor{blond}{rgb}{0.98, 0.94, 0.75}

\def \ie {\emph{i.e.}}
\def \eg {\emph{e.g.}}
\def \etal {\emph{et al.}}

\newcommand{\tit}[1]{\smallbreak\noindent\textbf{#1.}}
\newcommand{\tinytit}[1]{\noindent\textbf{#1.}}

\newcommand{\cmark}{\ding{51}}%
\newcommand{\xmark}{\ding{55}}%

\newcommand{\ours}{MGD\xspace}
\newcommand{\dataset}{Dress Code Multimodal\xspace}
\newcommand{\datasetviton}{VITON-HD Multimodal\xspace}

\newcommand\blfootnote[1]{%
  \begingroup
  \renewcommand\thefootnote{}\footnote{#1}%
  \addtocounter{footnote}{-1}%
  \endgroup
}

\iccvfinalcopy 

\def\iccvPaperID{7336} 

\begin{document}

\title{Multimodal Garment Designer:\\Human-Centric Latent Diffusion Models for Fashion Image Editing}

\author{Alberto Baldrati$^{1,3,*}$, Davide Morelli$^{2,3,*}$, Giuseppe Cartella$^2$, Marcella Cornia$^2$,\\Marco Bertini$^1$, Rita Cucchiara$^{2,4}$ \\
$^1$University of Florence, Italy \quad $^2$University of Modena and Reggio Emilia, Italy \\ $^3$University of Pisa, Italy \quad $^4$IIT-CNR, Italy\\
{\tt\small $^1$\{name.surname\}@unifi.it} \quad\quad {\tt\small $^2$\{name.surname\}@unimore.it}
}


\begin{figure}[htb]
\twocolumn[{
\renewcommand\twocolumn[1][]{#1}%
\maketitle
\vspace{-25pt}
\begin{center}
\includegraphics[width=0.995\linewidth]{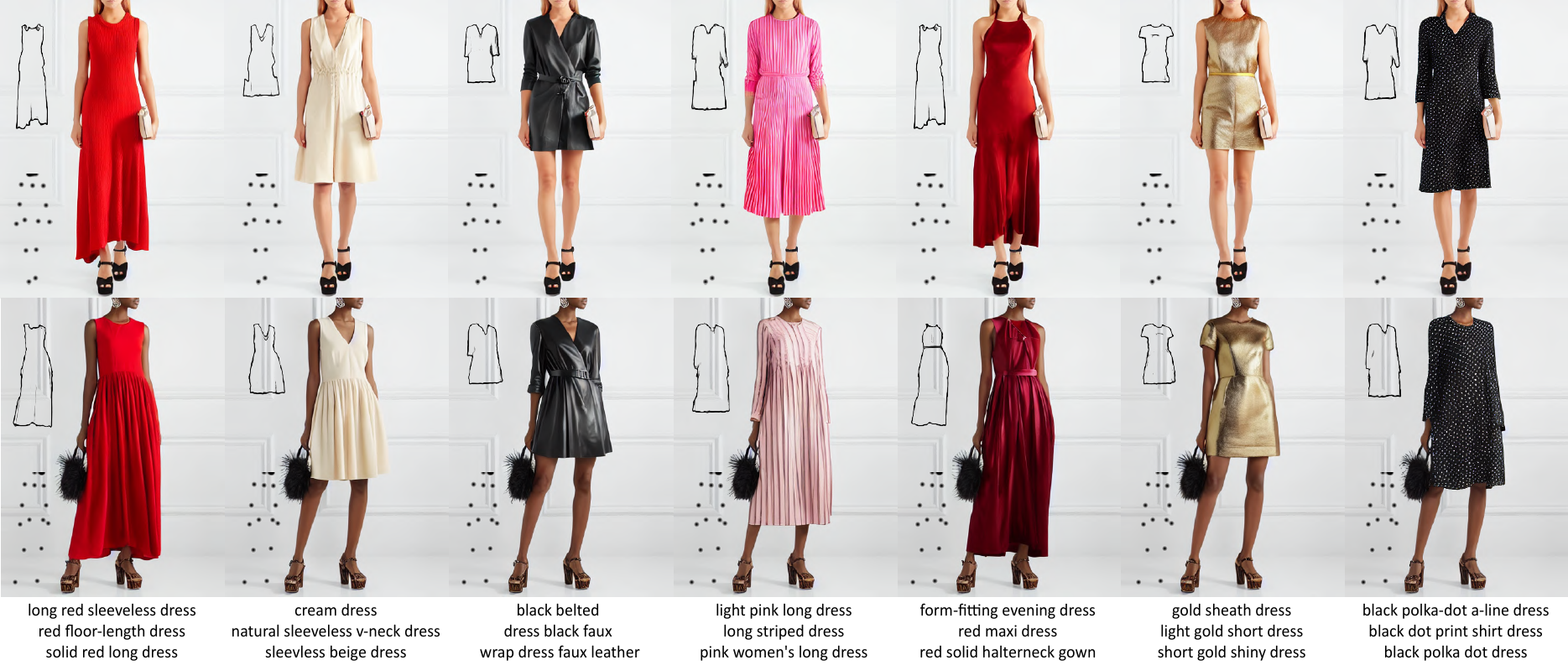}
\end{center}
\vspace{-0.5cm}
\caption{In this work, we propose a novel multimodal garment designer framework based on latent diffusion models that can generate a novel fashion image conditioned on text, human keypoints, and a garment sketch.}
\label{fig:first_page}
\vspace{0.3cm}
}]
\end{figure}

\vspace{-22pt}
\begin{abstract}
\vspace{-9pt}
Fashion illustration is used by designers to communicate their vision and to bring the design idea from conceptualization to realization, showing how clothes interact with the human body. In this context, computer vision can thus be used to improve the fashion design process. Differently from previous works that mainly focused on the virtual try-on of garments, we propose the task of multimodal-conditioned fashion image editing, guiding the generation of human-centric fashion images by following multimodal prompts, such as text, human body poses, and garment sketches. We tackle this problem by proposing a new architecture based on latent diffusion models, an approach that has not been used before in the fashion domain. Given the lack of existing datasets suitable for the task, we also extend two existing fashion datasets, namely Dress Code and VITON-HD, with multimodal annotations collected in a semi-automatic manner. Experimental results on these new datasets demonstrate the effectiveness of our proposal, both in terms of realism and coherence with the given multimodal inputs. Source code and collected multimodal annotations are publicly available at: \small{\href{https://github.com/aimagelab/multimodal-garment-designer}{https://github.com/aimagelab/multimodal-garment-designer}}.
\blfootnote{$^*$Equal contribution.}
\vspace{-6pt}
\end{abstract}

\section{Introduction}
\label{sec:intro}
Computer Vision research has always paid much attention both to the human person and to fashion-related problems, especially working on the recognition and retrieval of clothing items~\cite{hadi2015buy,liu2016deepfashion}, the recommendation of similar garments~\cite{hsiao2018creating,cucurull2019context,sarkar2023outfittransformer}, and the virtual try-on of clothes and accessories~\cite{han2018viton,wang2018toward,yang2020towards,morelli2022dresscode,choi2021viton,morelli2023ladi}. In the last years, some research efforts have been dedicated to the text-conditioned image editing task where, given a model image and a textual description of a garment, the goal is to generate the input model wearing a new clothing item corresponding to the given textual description. In this context, only a few works~\cite{zhu2017your,jiang2022text2human,pernuvs2023fice} have been proposed, exclusively employing GAN-based approaches for the generative step.

Recently, diffusion models~\cite{ho2020denoising,dhariwal2021diffusion,nichol2021improved,sohl2015deep} have attracted more and more attention due to their outstanding generation capabilities, allowing the improvement of a variety of downstream tasks in several domains, while their applicability to the fashion domain is still unexplored. Many different solutions have been introduced and can roughly be identified based on the denoising conditions used to guide the diffusion process, which can enable greater control of the synthesized output. A particular type of diffusion model has been proposed in~\cite{rombach2022high} that, instead of applying the diffusion process in the pixel space, defines the forward and the reverse processes in the latent space of a pre-trained autoencoder, becoming one of the leading choices thanks to its reduced computational cost. Although this solution can generate highly realistic images, it does not perform well in human-centric generation tasks and can not deal with multiple conditioning signals to guide the generation phase.

In this work, we address an extended and more general framework and define the new task of \textit{multimodal-conditioned fashion image editing}, which allows guiding the generative process via multimodal prompts while preserving the identity and body shape of a given person (Fig.~\ref{fig:first_page}). To tackle this task, we introduce a new architecture, called Multimodal Garment Designer (\ours), that emulates the process of a designer conceiving a new garment on a model shape, based on preliminary indications provided through a textual sentence or a garment sketch. In particular, starting from Stable Diffusion~\cite{rombach2022high}, we propose a denoising network that can be conditioned by multiple modalities and also takes into account the pose consistency between input and generated images, thus improving the effectiveness of human-centric diffusion models.

To address the newly proposed task, we present a semi-automatic framework to extend existing datasets with multimodal data. Specifically, we start from two famous virtual try-on datasets (\ie~Dress Code~\cite{morelli2022dresscode} and VITON-HD~\cite{choi2021viton}) and extend them with textual descriptions and garment sketches. Experimental results on the two proposed multimodal fashion benchmarks show both quantitatively and qualitatively that our proposed architecture generates high-quality images based on the given multimodal inputs and outperforms all considered competitors and baselines, also according to human evaluations.

To sum up, our contributions are as follows: (1) We propose a novel task of multimodal-conditioned fashion image editing, which entails the use of multimodal data to guide the generation. (2) We introduce a new human-centric generative architecture based on latent diffusion models, capable of following multimodal prompts while preserving the model's characteristics. (3) To tackle the new task, we extend two existing fashion datasets with textual sentences and garment sketches devising a semi-automatic annotation framework. (4) Extensive experiments demonstrate that the proposed approach outperforms other competitors in terms of realism and coherence with multimodal inputs.

\section{Related Work}
\label{sec:related}
\tinytit{Text-Guided Image Generation}
Creating an image that faithfully reflects the provided textual prompt is the goal of text-to-image synthesis. In this context, early approaches were based on GANs~\cite{xu2018attngan,zhu2019dm,zhang2021cross,tao2022df}, while most recent solutions exploit the effectiveness of diffusion models~\cite{nichol2022glide,ramesh2022hierarchical,rombach2022high}. In the fashion domain, only a few attempts of text-to-image synthesis have been proposed~\cite{zhu2017your,jiang2022text2human,pernuvs2023fice}. Specifically, Zhu~\etal~\cite{zhu2017your} presented a GAN-based solution that generates the final image conditioned on both textual descriptions and semantic layouts. A different approach is the one introduced in~\cite{pernuvs2023fice}, where a latent code regularization technique is employed to augment the GAN inversion process by exploiting CLIP textual embeddings~\cite{Radford2021LearningTV} to guide the image editing process. Instead, Jiang~\etal~\cite{jiang2022text2human} proposed an architecture that synthesizes full-body images by mapping the textual descriptions of clothing items into one-hot vectors, limiting however the expressiveness capability of the conditioning signal.

\begin{figure*}[t]
\begin{center}
\includegraphics[width=\linewidth]{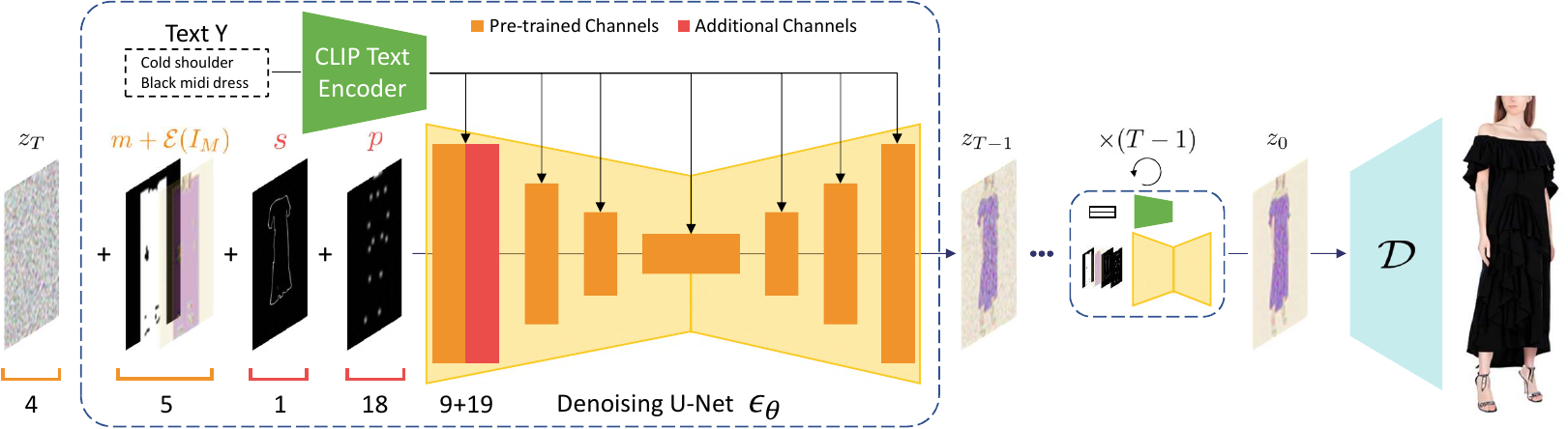}
\end{center}
\vspace{-0.4cm}
\caption{Overview of the proposed Multimodal Garment Designer (MGD), a human-centric latent diffusion model conditioned on multiple modalities (\ie~text, human pose, and garment sketch).} 
\label{fig:model}
\vspace{-0.3cm}
\end{figure*}

\tit{Multimodal Image Generation with Diffusion Models}
A related line of works aims to condition existing diffusion models on different modalities thus enabling greater control over the generation process~\cite{choi2021ilvr,meng2022sdedit,wang2022pretraining,mou2023t2i,cheng2023adaptively}. For example, Choi~\etal~\cite{choi2021ilvr} proposed to refine the generative process of an unconditional denoising diffusion probabilistic model~\cite{nichol2021improved} by matching each latent variable with the given reference image. On a different line, the approach introduced in~\cite{meng2022sdedit} adds noise to a stroke-based input and applies the reverse stochastic differential equation to synthesize images, without additional training. Wang~\etal~\cite{wang2022pretraining}, instead, proposed to learn a highly semantic latent space and perform conditional finetuning for each downstream task to map the guidance signals to the pre-trained space. 
Other recent works proposed to add sketches as additional conditioning signals, either concatenating them with the model input~\cite{cheng2023adaptively} or training an MLP-based edge predictor to map latent features to spatial maps~\cite{voynov2022sketch}.

Among contemporary works that aim to condition pre-trained latent diffusion models, ControlNet~\cite{zhang2023adding} proposes to extend the Stable Diffusion model~\cite{rombach2022high} with an additional conditioning input. This process involves creating two versions of the original model's weights: one that remains fixed and unchanged (locked copy) and another that can be updated during training (trainable copy). The purpose of this is to allow the trainable version to learn the newly introduced condition while the locked version retains the original model knowledge. On the other hand, T2I-Adapter~\cite{mou2023t2i} learns modality-specific adapter modules that enable Stable Diffusion conditioning on new modalities.

In contrast, we focus on the fashion domain and propose a human-centric architecture based on latent diffusion models that directly exploits the conditioning of textual sentences and other modalities such as human body poses and garment sketches.

\section{Proposed Method}
\label{sec:method}
In this section, we propose a novel task to automatically edit a human-centric fashion image conditioned on multiple modalities.
Specifically, given the model image $I \in \mathbb{R}^{H \times W \times 3}$, its pose map $P \in \mathbb{R}^{H \times W \times 18}$ where the channels represent the human keypoints, a textual description $Y$ of a garment, and a sketch of the same $S \in \mathbb{R}^{H \times W \times 1}$, we want to generate a new image $\Tilde{I} \in \mathbb{R}^{H \times W \times 3}$ that retains the information of the input model while substituting the target garment according to the multimodal inputs.
To tackle the task, we propose a novel latent diffusion approach, called Multimodal Garment Designer (MGD), that can effectively combine multimodal information when generating the new image $\Tilde{I}$.
Our proposed architecture is a general framework that can be easily extended to other modalities such as texture and 3d information. We strongly believe this task can foster research in the field and enhance the design process of new fashion items with greater customization. An overview of our model is shown in Fig.~\ref{fig:model}.

\subsection{Preliminaries}

While diffusion models~\cite{sohl2015deep} are latent variable architectures that work in the same dimensionality of the data (\ie~in the pixel space), latent diffusion models (LDMs)~\cite{rombach2022high} operate in the latent space of a pre-trained autoencoder achieving higher computational efficiency while preserving the generation quality. In our work, we leverage the Stable Diffusion model~\cite{rombach2022high}, a text-to-image implementation of LDMs as a starting point to perform multimodal conditioning for human-centric fashion image editing. Stable Diffusion is composed of an autoencoder with an encoder $\mathcal{E}$ and a decoder $\mathcal{D}$, a text-time-conditional U-Net denoising model $\epsilon_{\theta}$, and a CLIP-based text encoder $T_E$ taking as input a text $Y$.
The encoder $\mathcal{E}$ compresses an image $I$ into a lower-dimensional latent space defined in $\mathbb{R}^{h \times w \times 4}$, where $h=H/8$ and $w=W/8$. The decoder $\mathcal{D}$ performs the opposite operation, decoding a latent variable into the pixel space. For the sake of clarity, we define the $\epsilon_{\theta}$ convolutional input (\ie~$z_t$ in this case) as spatial input $\gamma$ because of the property of convolutions to preserve the spatial structure and the attention conditioning input as $\psi$.
The denoising network $\epsilon_{\theta}$ is trained according to the following loss:
\begin{equation}
    L = \mathbb{E}_{\mathcal{E}(I), Y, \epsilon \sim \mathcal{N}(0,1),t} \left[ \lVert \epsilon - \epsilon_{\theta}(\gamma,\psi) \rVert_2^2 \right],
    \label{eq:diffusion_loss}
\end{equation}
where $t$ is the diffusing time step, $\gamma = z_t$, $\psi=\left[t;T_E(Y)\right]$, and $\epsilon \sim \mathcal{N}(0,1)$ is the Gaussian noise added to $\mathcal{E}(I)$.

\subsection{Human-Centric Image Editing}

Our task aims to generate a new image $\Tilde{I}$,
by replacing in the input image $I$ the target garment using multimodal inputs, while preserving the model's identity and physical characteristics. As a natural consequence, this task can be identified as a particular type of inpainting tailored for human body data. Instead of using a standard text-to-image model, we perform inpainting concatenating along the channel dimension of the denoising network input $z_t$ an encoded masked image $\mathcal{E}({I_M})$ and the relative resized binary inpainting mask $m \in \{0,1\}^{h \times w \times 1}$, which is derived from the original inpainting mask $M \in \{0,1\}^{H \times W \times 1}$. Since here, the spatial input of the denoising network is $\gamma = [z_t; m; \mathcal{E}(I_M)], \gamma \in \mathbb{R}^{h \times w \times 9}$. Thanks to the fully convolutional nature of the encoder $\mathcal{E}$ and the decoder $\mathcal{D}$, this LDMs-based architecture can preserve the spatial information in the latent space. Exploiting this feature, our method can thus optionally add conditioning constraints to the generation. In particular, we propose to add two generation constraints in addition to the textual information: the model pose map $P$ to preserve the original human pose of the input model and the garment sketch $S$ to allow the final users to better condition the garment generation process.

\tit{Pose Map Conditioning}
In most cases~\cite{suvorov2022resolution,lugmayr2022repaint,li2022mat}, inpainting is performed with the objective of either removing or entirely replacing the content of the masked region. However, in our task, we aim to remove all information regarding the garment worn by the model while preserving the model's body information and identity. Thus, we propose to improve the garment inpainting process by using the bounding box of the segmentation mask along with pose map information representing body keypoints. This approach enables the preservation of the model's physical characteristics in the masked region while allowing the inpainting of garments with different shapes. Differently from conventional inpainting techniques, we focus on selectively retaining and discarding specific information within the masked region to achieve the desired outcome.
To enhance the performance of the denoising network with human body keypoints, we modify the first convolution layer of the network by adding 18 additional channels, one for each keypoint. Adding new inputs usually would require retraining the model from scratch, thus consuming time, data, and resources, especially in the case of data-hungry models like the diffusion ones. Therefore, we propose to extend the kernels of the pre-trained input layer of the denoising network with randomly initialized weights sampled from a uniform distribution~\cite{he2015delving} and retrain the whole network. This consistently reduces the number of training steps and enables training with less data. Our experiments show that such improvement enhances the consistency of the body information between the generated image and the original one.

\tit{Incorporating Sketches}
Fully describing a garment using only textual descriptions is a challenging task due to the complexity and ambiguity of natural language. While text can convey specific attributes like style, color, and patterns of a garment, it may not provide sufficient information about its spatial characteristics, such as shape and size. This limitation can hinder the customization of the generated clothing item other than the ability to accurately match the user's intended style. Therefore, we propose to leverage garment sketches to enrich the textual input with additional spatial fine-grained details. We achieve this following the same approach described for pose map conditioning.
The final spatial input of our denoising network is $\gamma = \left[z_t; m; \mathcal{E}(I_M); p;s\right]$, $\left[p; s\right] \in \mathbb{R}^{h \times w \times (18+1)}$, $p$ and $s$ are obtained by resizing $P$ and $S$ to match the latent space dimensions. In the case of sketches, we only condition the early steps of the denoising process as the final steps have little influence on the shapes~\cite{balaji2022ediffi}.

\tit{Mask Composition}
To preserve the model identity when performing human-centric inpainting, we perform mask composition as the final step of the proposed approach.
Defining $\hat{I} = \mathcal{D}(z_0) \in \mathbb{R}^{H \times W \times 3}$ as the output of the decoder $\mathcal{D}$ and $M_{\text{head}} \in \{0,1\}^{H \times W \times 1}$ as the model face binary mask of the image $I$,
the final output image $\Tilde{I}$ is obtained as follows:
$\Tilde{I} = M_{\text{head}}\odot I + (1-M_{\text{head}})\odot \hat{I}$, 
where $\odot$ denotes the element-wise multiplication operator.

\subsection{Training and Inference}

As in standard latent diffusion models, given an encoded input $z = \mathcal{E}(I)$, the proposed denoising network is trained to predict the noise stochastically added to $z$. The corresponding objective function can be specified as

\begin{equation}
\label{eq:our_loss}
    L = \mathbb{E}_{\mathcal{E}(I), Y, \epsilon \sim \mathcal{N}(0,1), t, \mathcal{E}(I_M),m,p,s} \left[ \lVert \epsilon - \epsilon_{\theta}(\gamma,\psi) \rVert_2^2 \right],
\end{equation}
where $\gamma=\left[z_t; m; \mathcal{E}(I_M); p; s\right]$ and $\psi=\left[t; T_E(Y)\right]$.

\tit{Classifier-Free Guidance}
Classifier-free guidance is an inference technique 
that requires the denoising network to work both conditioned and unconditioned. This method modifies the unconditional model predicted noise moving it toward the conditioned one.
Specifically, the predicted diffusion process at time $t$, given the generic condition $c$, is computed as follows:
\begin{equation}
    \hat{\epsilon}_{\theta}(z_t | c) = \epsilon_{\theta}(z_t | \emptyset) + \alpha \cdot (\epsilon_{\theta}(z_t | c) - \epsilon_{\theta}(z_t | \emptyset)),
    \label{eq:diffusion_classifier_free}
\end{equation}
where $\epsilon_{\theta}(z_t | c)$ is the predicted noise at time $t$ given the condition $c$, $\epsilon_{\theta}(z_t | \emptyset)$ is the predicted noise at time $t$ given the null condition, and the guidance scale $\alpha$ controls the degree of extrapolation towards the condition.

Since our model deals with three conditions (\ie~text, pose map, and sketch), we use the fast variant multi-condition classifier-free guidance proposed in~\cite{avrahami2022spatext}. 
Instead of performing the classifier-free guidance according to each condition probability, it computes the direction of the joint probability of all the conditions $\Delta_{\text{joint}}^t = \epsilon_{\theta}(z_t | \{ c_i \}_{i=1}^{i=N}) - \epsilon_{\theta}(z_t | \emptyset)$:
\begin{equation}
    \hat{\epsilon}_{\theta}(z_t | \{ c_i \}_{i=1}^{i=N}) = \epsilon_{\theta}(z_t | \emptyset) + \alpha \cdot \Delta_{\text{joint}}^t. 
    \label{eq:diffusion_classifier_free2}
\end{equation}
This reduces the number of feed-forward executions from $N+1$ to $2$.

\tit{Unconditional Training}
Ensuring the ability of the denoising model to work both with and without conditions is achieved by replacing at training time the condition with a null one according to a fixed probability. 
This approach allows the model to learn from both conditional and unconditional samples, resulting in improved mode coverage and sample fidelity. Moreover, this technique also allows the model to optionally use the control signals at prediction time. Since our approach considers multiple conditions, we propose to extend the input masking to each condition independently. Experiments show that tuning this parameter can effectively affect the quality of the final result.

\section{Collecting Multimodal Fashion Datasets}
\label{sec:dataset}
\begin{figure}[t]
\begin{center}
    \includegraphics[width=\linewidth]{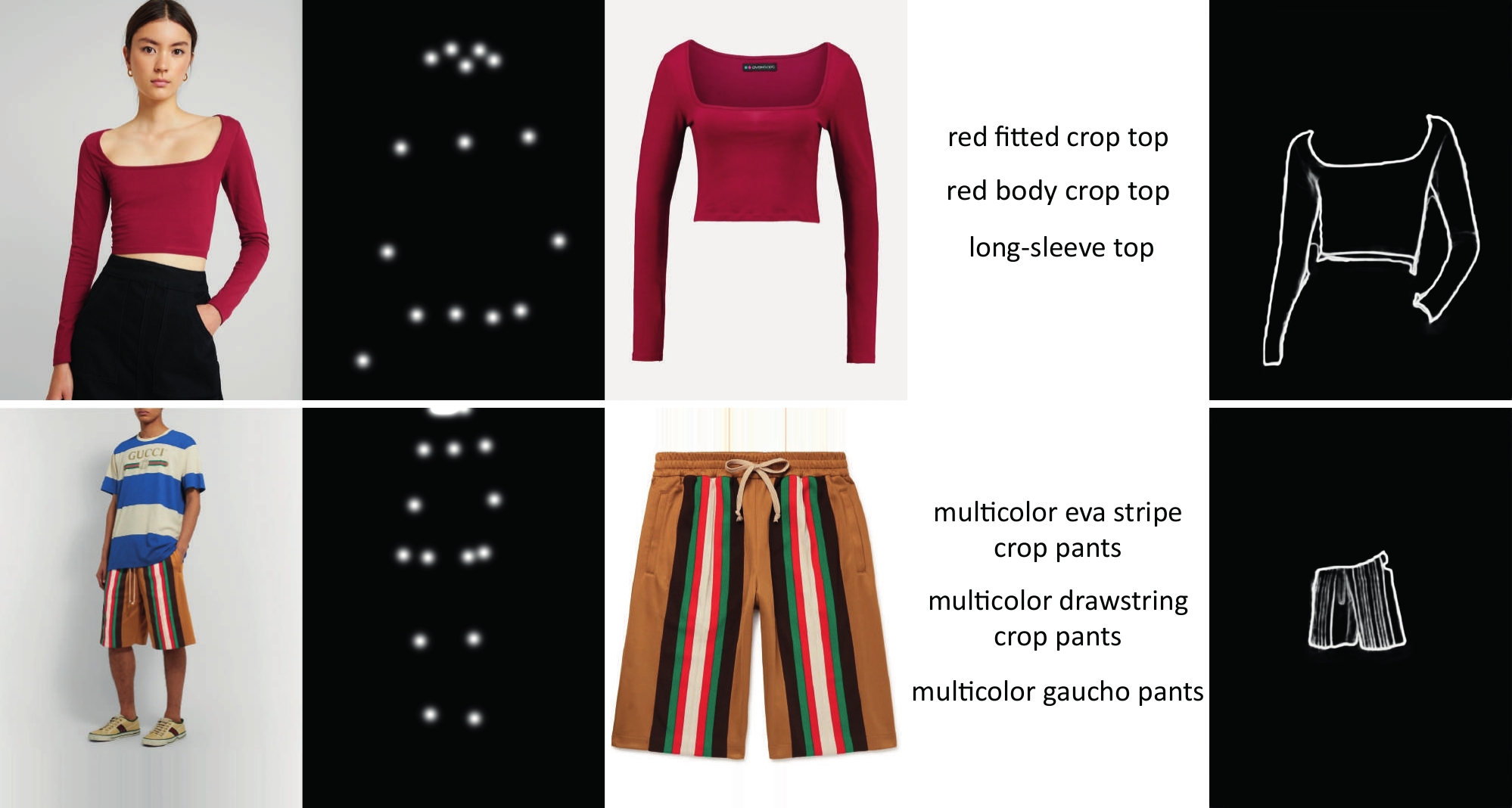}
\end{center}    
\vspace{-0.4cm}
\caption{Sample images and multimodal data from our newly collected datasets.}
\label{fig:dataset}
\vspace{-.15cm}
\end{figure}

Currently available datasets for fashion image generation often contain low-resolution images and lack all the required multimodal information needed to perform the task previously described. For this reason, the collection of new multimodal datasets for the fashion domain plays a crucial role to advance research in the field. To this aim, we start from two recent high-resolution fashion datasets introduced for the virtual try-on task, namely Dress Code~\cite{morelli2022dresscode} and VITON-HD~\cite{choi2021viton}, and extend them with textual sentences and garment sketches. Both datasets include image pairs with a resolution of $1024\times768$, each composed of a garment image and a reference model wearing the given fashion item. In this section, we introduce a framework to semi-automatically annotate fashion images with multimodal information and provide a complete description of how to enrich Dress Code and VITON-HD with garment-related text and sketches. We call our extended versions of these datasets \dataset and \datasetviton, respectively. Sample images and multimodal data of the collected datasets can be found in Fig.~\ref{fig:dataset}.

\subsection{Dataset Collection and Annotation}

\tinytit{Data Preparation}
We start the annotation from the Dress Code dataset, which contains more than 53k model-garment pairs of multiple categories. As a first step, we need to associate each garment with a textual description containing fashion-specific and non-generic terms which are sufficiently detailed but not extremely lengthy to be exploited for constraining the generation. Motivated by recent findings in the field showing that humans tend to describe fashion items using only a few words~\cite{bianchi2021query2prod2vec}, we propose to use noun chunks (\ie~short textual sentences composed of a noun along with its modifiers) that can effectively capture important information while reducing unnecessary words or details.
Given that manually annotating all the images would be time-consuming and resource-intensive\footnote{Since the Dress Code dataset consists of over 53k fashion items and assuming that each annotation requires approximately 5 minutes, a single annotator working 8 hours per day, 5 days a week, and 260 working days per year would take more than 2 years to complete the annotation task.}, we propose a novel framework to semi-automatically annotate the dataset using noun chunks. Firstly, domain-specific captions are collected from two available fashion datasets, namely FashionIQ~\cite{wu2021fashion} and Fashion200k~\cite{han2017automatic}, standardizing them with word lemmatization and eventually reducing each word to its root form with the NLTK library\footnote{\href{https://www.nltk.org/}{https://www.nltk.org/}}. Then, we extract noun chunks from the captions, filtering the results by removing all textual items that start with or contain special characters. After this pre-processing stage, we obtain more than 60k unique noun chunks, divided into three different categories (\ie~upper-body clothes, lower-body clothes, and dresses).

To determine the most relevant noun chunks for each garment, we employ the CLIP model~\cite{Radford2021LearningTV} and its open-source adaptation (\ie~OpenCLIP~\cite{wortsman2022robust}). We select the VIT-L14@336 and RN50$\times$64 models for CLIP, and the VIT-L14, ViT-H14, and ViT-g14 models for OpenCLIP. Prompt ensembling is performed to improve the results and, for each image, we select 25 noun chunks based on the top-5 noun chunks per model rated by cosine similarity between image and text embeddings, avoiding repetitions. 

\tit{Fine-Grained Textual Annotation}
To ensure the accuracy and representativeness of our annotations, we manually annotate a significant portion of Dress Code images. In particular, we select the three most representative noun chunks, among the 25 automatically associated, with each garment image. To minimize the annotation time, we develop a custom annotation tool that constrains the annotation time to an average time of 60 seconds per item and allows the annotator to manually insert noun chunks in the case that none of the automatically extracted ones are suitable for the image. Overall, we manually annotate 26,400 different garments (8,800 for each category) out of the 53,792 products included in the dataset, ensuring to include all fashion items of the original test set~\cite{morelli2022dresscode}.

\begin{table}[t]
\begin{center}
\footnotesize
\setlength{\tabcolsep}{.3em}
\resizebox{\linewidth}{!}{
\begin{tabular}{lc cc cc cc}
\toprule
& & & & & & \textbf{\# Unique} & \textbf{\# Unique} \\
\textbf{Dataset} & \textbf{Text} & \textbf{Pose} & \textbf{Sketch} & \textbf{\# Images} & \textbf{\# Products} & \textbf{Texts} & \textbf{Words} \\
\midrule
VITON-HD~\cite{choi2021viton} & \xmark & \cmark & \xmark & 27,358 & 13,679 & - & - \\
Dress Code~\cite{morelli2022dresscode} & \xmark & \cmark & \xmark & 107,584 & 53,792 & - & - \\
\midrule
Be Your Own Prada~\cite{zhu2017your} & \cmark & \cmark & \xmark & 78,979 & N/A & 3,972 & 445 \\
DF-Multimodal~\cite{jiang2022text2human} & \cmark & \cmark & \xmark & 44,096 & N/A & 10,253 & 77 \\
\midrule
\textbf{\datasetviton} & \cmark & \cmark & \cmark & 27,358 & 13,679 & 5,143 & 1,613 \\
\textbf{\dataset} & \cmark & \cmark & \cmark & 107,584 & 53,792 & 25,596 & 2,995 \\
\bottomrule
\end{tabular}
}
\end{center}
\vspace{-0.4cm}
\caption{Comparison of Dress Code and \datasetviton with other fashion datasets with multimodal annotations.}
\label{tab:datasets_comparison}
\vspace{-0.35cm}
\end{table}

\tit{Coarse-Grained Textual Annotation}
To complete the annotation, we first finetune the OpenCLIP ViT-B32 model, pre-trained on the English portion of the LAION5B dataset~\cite{schuhmann2022laionb}, using the newly annotated image-text pairs. We then use this model and the collected set of noun chunks to automatically tag all the remaining elements of the Dress Code dataset with the three most similar noun chunks, always determined via cosine similarity between multimodal embeddings.
We employ the same strategy also to automatically annotate all garment images of the VITON-HD dataset. In this case, since this dataset only contains upper-body clothes, we limit the table noun chunks to the ones describing upper-body garments. 

\tit{Extracting Sketches} The introduction of garment sketches can provide valuable design details that are not easily discernible from text alone. In this way, the dataset can provide a more accurate and comprehensive representation of the garments, leading to improved quality and better control of the generated design details. To extract sketches for both Dress Code and VITON-HD datasets, we employ PiDiNet~\cite{su2021pixel}, a pre-trained edge detection network. 

Given that the selected datasets have originally been introduced for virtual try-on, they consist of both paired and unpaired test sets. While for the paired set we can directly use the human parsing mask to extract the garment of interest worn by the model and then feed it to the edge detection network, for the unpaired set we need to first create a warped version of the in-shop garment matching the body pose and shape of the target model. Following virtual try-on methods~\cite{wang2018toward,yang2020towards}, we train a geometric transformation module that performs a thin-plate spline transformation~\cite{rocco2017convolutional} of the input garment and then refines the warped result using a U-Net model~\cite{ronneberger2015u}. From each warped garment, we extract the sketch image enabling the use of the proposed solution even in unpaired settings.

\subsection{Comparison with Other Datasets}

The only two text-to-image generation datasets available in the fashion domain~\cite{zhu2017your,jiang2022text2human} are both based on images from the DeepFashion dataset~\cite{liu2016deepfashion}.
While the dataset introduced in~\cite{zhu2017your} contains short textual descriptions, DeepFashion-Multimodal~\cite{jiang2022text2human} is annotated with attributes (\eg~category, color, fabric, etc.) that can be composed in longer captions. In Table~\ref{tab:datasets_comparison}, we summarize the main statistics of the publicly available datasets textual annotations compared with those of our newly extended datasets. As can be seen, our datasets contain more variety in terms of textual items and words, confirming the appropriateness of our annotation procedure and enabling a more personalized control of the generation process. Also, it is worth noting that the other datasets have no in-shop garment images making them difficult to employ in our case.

\section{Experimental Evaluation}
\label{sec:experiments}
\subsection{Implementation Details and Competitors}
\tit{Training and Inference}
All models are trained on the original splits of the \dataset and \datasetviton datasets on a single NVIDIA A100 GPU for 150k steps, using a batch size of 16, a learning rate of $10^{-5}$ with a linear warmup for the first 500 iterations, and AdamW~\cite{loshchilov2019decoupled} as optimizer with weight decay $10^{-2}$. To speed up training and save memory, we use mixed precision~\cite{micikevicius2018mixed}. We set both the fraction of steps conditioned by the sketch and the portion of masked conditions during training to $0.2$. 
During inference, we employ the DDIM~\cite{song2021denoising} with 50 steps as noise scheduler and set the classifier-free guidance parameter $\alpha$ to 7.5.

\begin{table*}[ht!]
\begin{center}
\footnotesize
\setlength{\tabcolsep}{.4em}
\resizebox{\linewidth}{!}{
\begin{tabular}{lcc ccc c ccccc c ccccc}
\toprule
 & & & \multicolumn{3}{c}{\textbf{Modalities}} & & \multicolumn{5}{c}{\textbf{\dataset}}  & & \multicolumn{5}{c}{\textbf{\datasetviton}} \\
\cmidrule{4-6} \cmidrule{8-12} \cmidrule{14-18}
\textbf{Model} & \textbf{Resolution} & & \textbf{Text} & \textbf{Pose} & \textbf{Sketch} & & \textbf{FID} $\downarrow$ & \textbf{KID} $\downarrow$ & \textbf{CLIP-S} $\uparrow$ & \textbf{PD} $\downarrow$ & \textbf{SD} $\downarrow$ & & \textbf{FID} $\downarrow$ & \textbf{KID} $\downarrow$ & \textbf{CLIP-S} $\uparrow$ & \textbf{PD} $\downarrow$ & \textbf{SD} $\downarrow$ \\
\midrule
\textit{Paired setting} \\
\hspace{0.4cm}Stable Diffusion~\cite{rombach2022high} & 256$\times$192 & & \cmark & & & & 17.05 & 9.28 & 28.71 & 4.62 & - & & 15.18 & 6.38 & 30.40 & 5.04 & - \\
\hspace{0.4cm}FICE~\cite{pernuvs2023fice} & 256$\times$192 & & \cmark & \cmark & & & 30.63 & 23.54 & 28.72 & 6.87 & - & & 49.44 & 44.74 & 29.26 & 6.37 & - \\
\rowcolor{blond}
\hspace{0.4cm}\textbf{\ours (ours)} & 256$\times$192 & & \cmark & \cmark & & & \bf5.57 & \bf1.67 & \bf31.33 & \bf2.37 & - & & \textbf{10.11} & \bf3.14 & \bf31.85 & \bf2.90 & - \\
\midrule
\textit{Paired setting} \\
\hspace{0.4cm}Stable Diffusion~\cite{rombach2022high} & 512$\times$384 & & \cmark & & & & 17.43 & 9.48 & 29.18 & 9.24 & 0.467 & & 16.28 & 6.56 & 30.70 & 10.78 & 0.410 \\
\hspace{0.4cm}SDEdit~\cite{meng2022sdedit} & 512$\times$384 & & \cmark & \cmark & \cmark & & 10.19 & 5.03 & 29.21 & 5.41 & 0.398 & & 13.07 & 4.66 & 30.58 & 6.76 & 0.306 \\
\rowcolor{blond}
\hspace{0.4cm}\textbf{\ours (ours)} & 512$\times$384 & & \cmark & \cmark & \cmark & & \bf5.74 & \bf2.11 & \bf31.68 & \bf4.72 & \bf0.374 & & \bf10.60 & \bf3.26 & \bf32.39 & \bf5.94 & \bf0.253 \\
\midrule
\textit{Unpaired setting} \\
\hspace{0.4cm}Stable Diffusion~\cite{rombach2022high} & 256$\times$192 & & \cmark & & & & 19.11 & 10.69 & 27.53 & 5.07 & - & & 17.37 & 7.55 & 28.40 & 5.50 & - \\
\hspace{0.4cm}FICE~\cite{pernuvs2023fice} & 256$\times$192 & & \cmark & \cmark & & & 34.14 & 26.86 & 26.03 & 7.15 & - & & 52.74 & 48.58 & 25.94 & 6.58 & - \\
\rowcolor{blond}
\hspace{0.4cm}\textbf{\ours (ours)} & 256$\times$192 & & \cmark & \cmark & & & \bf7.01 & \bf2.19 & \bf29.58 & \bf2.96 & - & & \bf11.54 & \bf3.18 & \bf29.95 & \bf3.30 & - \\
\midrule
\textit{Unpaired setting} \\
\hspace{0.4cm}Stable Diffusion~\cite{rombach2022high} & 512$\times$384 & & \cmark & & & & 19.55 & 10.80 & 28.02 & 9.89 & 0.582 & & 18.45 & 7.87 & 28.74 & 11.60 & 0.561 \\
\hspace{0.4cm}SDEdit~\cite{meng2022sdedit} & 512$\times$384 & & \cmark & \cmark & \cmark & & 11.38 & 5.69 & 27.10 & \bf6.16 & 0.509 & & 15.12 & 5.67 & 28.61 & 7.35 & 0.406 \\
\rowcolor{blond}
\hspace{0.4cm}\textbf{\ours (ours)} & 512$\times$384 & & \cmark & \cmark & \cmark & & \bf7.73 & \bf2.82 & \bf30.04 & 6.79 & \bf0.458 & & \bf12.81 & \bf3.86 & \bf30.75 & \bf7.22 & \bf0.317 \\
\bottomrule
\end{tabular}
}
\end{center}
\vspace{-0.4cm}
\caption{Quantitative results on the \dataset and \datasetviton datasets for both paired and unpaired settings  .
}
\vspace{-.4cm}
\label{tab:main_merged}
\end{table*}

\tit{Baselines and Competitors} As first competitor, we use the out-of-the-box implementation of the inpainting Stable Diffusion pipeline\footnote{\href{https://huggingface.co/runwayml/stable-diffusion-inpainting}{https://huggingface.co/runwayml/stable-diffusion-inpainting}} provided by Huggingface. Moreover, we adapt two existing models, namely FICE~\cite{pernuvs2023fice} and SDEdit~\cite{meng2022sdedit}, to work on our setting. In particular, we retrain all main components of the FICE model on the newly collected datasets. We employ the same resolution used by the authors (\ie~$256\times256$), downsampling each image to $256 \times 192$ and applying padding to match the desired size (which is then removed during evaluation). To compare our model with a different conditioning strategy, we employ the approach proposed in~\cite{meng2022sdedit} using our model trained only with text and human poses as input modalities and perform the sketch guidance using as starting latent variable the sketch image with added random noise. Following the original paper instructions, we use 0.8 as the strength parameter.

\subsection{Evaluation Metrics}
\label{sec:metrics}

To assess the realism of generated images, we employ the Fréchet Inception Distance (FID)~\cite{heusel2017gans} and the Kernel Inception Distance (KID)~\cite{binkowski2018demystifying}. For both metrics, we adopt the implementation proposed in~\cite{parmar2022aliased}.
Instead, to evaluate the adherence of the image to the textual conditioning input, we employ the CLIP Score (CLIP-S)~\cite{hessel2021clipscore} provided in the TorchMetrics library~\cite{detlefsen2022torchmetrics}, using the OpenCLIP ViT-H/14 model as cross-modal architecture. We compute the score on the inpainted region of the generated output pasted on a $224 \times 224$ white background.

\tit{Pose Distance (PD)} 
We propose a novel pose distance metric that measures the coherence of human body poses between the generated image and the original one estimating the distance between the human keypoints extracted from the original and generated images.
Specifically, we employ the OpenPifPaf~\cite{kreiss2021openpifpaf} human pose estimation network and compute the $\ell_2$ distance between each pair of real-generated corresponding estimated keypoints. We only consider the keypoints involved in the generation (\ie~that falls in the mask $M$) and weigh each keypoint distance with the detector confidence to take into account any estimation errors.

\begin{table}[ht!]
\begin{center}
\footnotesize
\setlength{\tabcolsep}{.3em}
\resizebox{\linewidth}{!}{
\begin{tabular}{lc ccc c ccccc}
\toprule
& & \multicolumn{3}{c}{\textbf{Modalities}} & & \multicolumn{5}{c}{\textbf{\dataset}} \\
\cmidrule{3-5} \cmidrule{7-11} 
\textbf{Model} & & \textbf{Text} & \textbf{Pose} & \textbf{Sketch} & & 
\textbf{FID} $\downarrow$ & \textbf{KID} $\downarrow$ & \textbf{CLIP-S} $\uparrow$  & \textbf{PD} $\downarrow$ & \textbf{SD} $\downarrow$ \\
\midrule
 & & \cmark & & & & 6.19 & 2.15 & \textbf{31.79} & 6.16 & 0.411 \\
 & & \cmark & \cmark & & & 6.31 & 2.33 & 31.67 & 5.31 & 0.405 \\
\rowcolor{blond}
\textbf{\ours (ours)} & & \cmark & \cmark & \cmark & & \textbf{5.74} & \textbf{2.11} & 31.68 & \textbf{4.72} & \textbf{0.374} \\
\midrule
& & \multicolumn{3}{c}{\textbf{Modalities}} & & \multicolumn{5}{c}{\textbf{\datasetviton}} \\
\cmidrule{3-5} \cmidrule{7-11} 
\textbf{Model} & & \textbf{Text} & \textbf{Pose} & \textbf{Sketch} & & 
\textbf{FID} $\downarrow$ & \textbf{KID} $\downarrow$ & \textbf{CLIP-S} $\uparrow$  & \textbf{PD} $\downarrow$ & \textbf{SD} $\downarrow$ \\
\midrule
 & & \cmark & & & & 11.39 & 3.52 & 32.16 & 7.83 & 0.339 \\
 & & \cmark & \cmark & & & 11.07 & 3.36 & 32.27 & 6.77 & 0.318 \\
\rowcolor{blond}
\textbf{\ours (ours)} & & \cmark & \cmark & \cmark & & \textbf{10.60} & \textbf{3.26} & \textbf{32.39} & \textbf{5.94} & \textbf{0.253} \\
\bottomrule
\end{tabular}
}
\end{center}
\vspace{-0.4cm}
\caption{Performance analysis on the paired setting of both datasets as input modalities vary.}
\vspace{-.35cm}
\label{tab:modalities}
\end{table}

\tit{Sketch Distance (SD)} 
To quantify the adherence of the generated image to the sketch constraint, we propose a novel sketch distance metric. To compute the score, we extract the segmentation map of the original and generated garments using an off-the-shelf clothing segmentation network\footnote{\href{https://github.com/levindabhi/cloth-segmentation}{https://github.com/levindabhi/cloth-segmentation}}. We then use the segmented garment area to extract garment sketches using the PIDInet~\cite{su2021pixel} edge detector network. The final score is the mean squared error between these sketches, weighting the per-pixel results on the inverse pixel frequency of the activated pixels. More details about these proposed metrics can be found in the supplementary.

\subsection{Experimental Results}

\tit{Comparison with Other Methods}
We test our proposal for the paired and unpaired settings of the considered datasets. In the former, the conditions (\eg~text, sketch) refers to the garment the model is wearing, while in the latter, the in-shop garment differs from the worn one. In Table~\ref{tab:main_merged}, we report the quantitative results on \dataset and \datasetviton in comparison with the aforementioned competitors. As can be seen, the proposed \ours model consistently outperforms competitors, in terms of realism (\ie~FID and KID) and coherency with input modalities (\ie~CLIP-S, PD, and SD). In particular, when considering low-resolution results, we notice that FICE~\cite{pernuvs2023fice} can produce images fairly consistent with the text conditioning, albeit less realistic than other methods. While Stable Diffusion~\cite{rombach2022high} enhances image realism, it fails to preserve the input model's pose due to the lack of pose information in the inputs. It is noteworthy that in this case, we compare the results of our model only using text and pose map as conditioning since both considered competitors are not conditioned on sketches. For this reason, we do not report the results in terms of sketch distance for low-resolution images. 

\begin{figure*}[t]
\begin{center}
    \includegraphics[width=\linewidth]{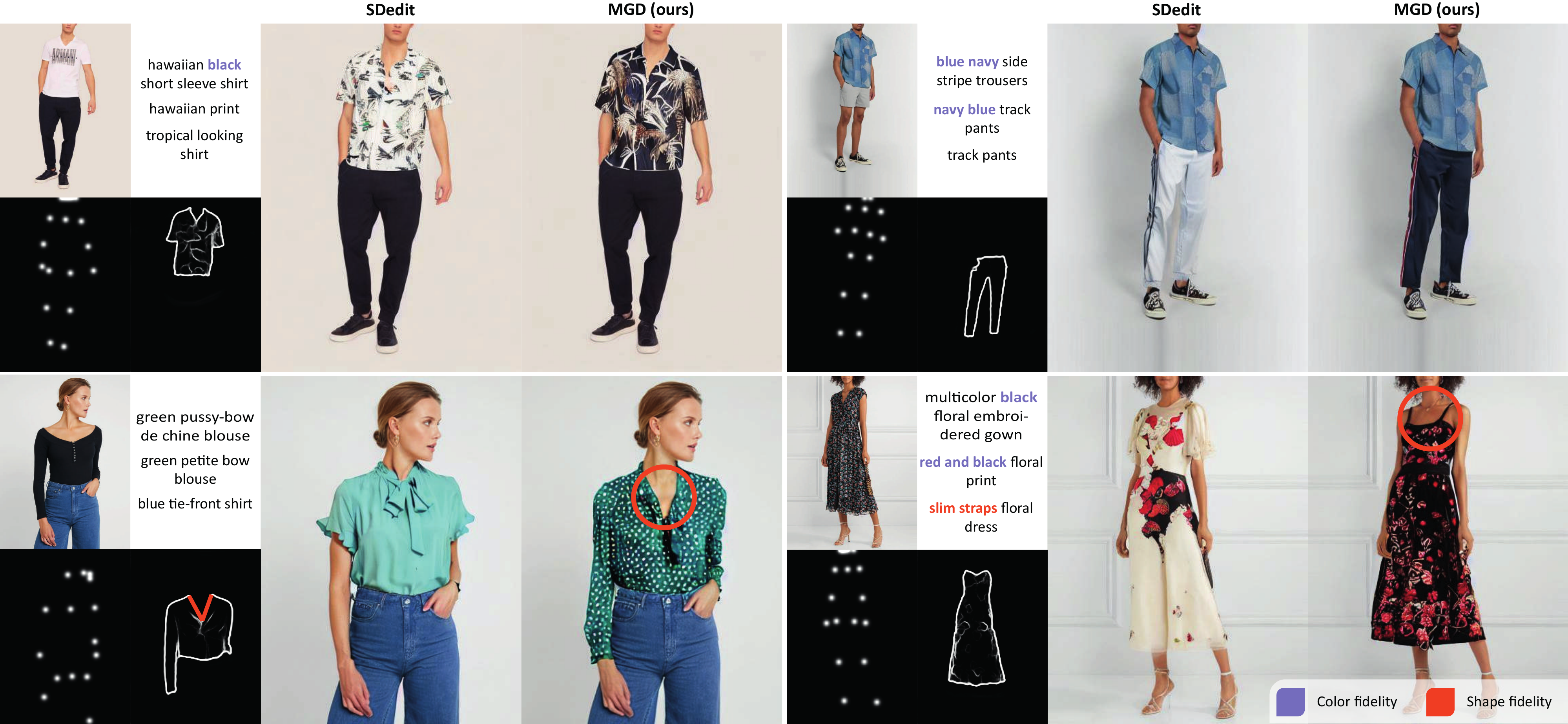}
\end{center}    
\vspace{-0.4cm}
\caption{Sample generated images on \dataset and \datasetviton (bottom left) using all multimodal inputs.}
\label{fig:qualitatives}
\vspace{-.4cm}
\end{figure*}

\begin{table}[t]
\begin{center}
\footnotesize
\setlength{\tabcolsep}{.32em}
\resizebox{\linewidth}{!}{
\begin{tabular}{cc ccc c ccc c ccc}
\toprule
& &\multicolumn{3}{c}{\textbf{Modalities}} & & \multicolumn{3}{c}{\textbf{Realism}} & & \multicolumn{3}{c}{\textbf{Multimodal Coherence}} \\
\cmidrule{3-5} \cmidrule{7-9} \cmidrule{11-13}
& & \textbf{Text} & \textbf{Pose} & \textbf{Sketch} & & 
\textbf{Stable Diff.} & \textbf{FICE} & \textbf{SDEdit} & & 
\textbf{Stable Diff.} & \textbf{FICE} & \textbf{SDEdit} \\
\midrule
\multirow{3}{*}{{\rotatebox[origin=c]{90}{Dress}}} & \multirow{3}{*}{{\rotatebox[origin=c]{90}{Code M.}}} & \cmark & & & & 70.82 & - & - & & 65.32 & - & - \\
& & \cmark & \cmark & & & 70.73 & 96.26 & - & & 65.15 & 84.48 & - \\
& & \cmark & \cmark & \cmark & & 70.29 & - & 52.54 & & 65.38 & - & 66.23 \\
\midrule
\multirow{3}{*}{{\rotatebox[origin=c]{90}{VITON}}} & \multirow{3}{*}{{\rotatebox[origin=c]{90}{HD M.}}} & \cmark & & & & 67.03 & - & - & & 57.76 & - & - \\
& & \cmark & \cmark & & & 66.17 & 93.84 & - & & 73.73 & 83.46 & -   \\
& & \cmark & \cmark & \cmark & & 60.71 & - & 53.44 & & 69.47 & - & 59.34 \\
\bottomrule
\end{tabular}
}
\end{center}
\vspace{-0.4cm}
\caption{User study results on the unpaired setting of both datasets. We report the percentage of times an image from \ours is preferred against a competitor. Comparisons with FICE~\cite{pernuvs2023fice} are performed at $256\times192$ resolution.}
\label{tab:user_study}
\vspace{-.35cm}
\end{table}

In the high-resolution setting, we evaluate instead our \ours method using all multimodal conditions (\ie~text, pose map, and sketch) as input. In this case, we compare \ours with Stable Diffusion~\cite{rombach2022high} plus SDEdit~\cite{meng2022sdedit}, where we use our text-pose conditioned denoising network as SDEdit backbone. Our findings indicate that Stable Diffusion performs worse in terms of the pose distance than both SDEdit and \ours, owing to the lack of pose information in the inputs. It is noteworthy that SDEdit performs worse than our model in all metrics. We attribute this behavior to the way sketch conditioning happens. In SDEdit, it occurs only at the beginning by initializing $z_t$ using the sketch image with added noise according to the conditioning strength, while our model conditions the denoising process in multiple steps, depending on the sketch conditioning parameter. 
Qualitative results reported in Fig.~\ref{fig:qualitatives} highlight how our model better follows the given conditions and generate high-realistic images. 

To validate our results based on human judgment, we conduct a user study that evaluates both the realism of the generation and the adherence to multimodal inputs. Overall we collect about 7k evaluations involving more than 150 users. Additional details are reported in the supplementary. Table~\ref{tab:user_study} shows the user study results. Also in this case our model outperforms the competitors, thus confirming the effectiveness of our proposal. 

\tit{Varying Input Modalities} 
In Table~\ref{tab:modalities}, we study the behavior of our \ours model when the input modalities are masked (\ie~where we feed the model with a zero tensor instead of the considered modality). In particular, we focus on the CLIP-S for text adherence and on the newly proposed pose and sketch distances for the pose and sketch coherency, respectively. Notice that the text input anchors the CLIP-S metrics of all experiments and makes them comparable in all cases. Starting from the fully conditioned model (\ie~text, pose, sketch), we mask the sketch. As the decrease of the sketch distance in Table~\ref{tab:modalities} confirms, this input actually influences the generation process of our model in both the considered datasets. Also, this modality slightly affects the pose distance as the sketch implicitly contains information about the model's body pose. We further mask the pose map input and compare the output with previous results. In this case, we can also notice a consistent difference with the text-only conditioned model, according to all metrics except CLIP-S as expected.
These results confirm that our \ours model can effectively deal with the conditions in a disentangled way, making them optional.

\begin{table}[t]
\begin{center}
\footnotesize
\setlength{\tabcolsep}{.35em}
\resizebox{\linewidth}{!}{
\begin{tabular}{cc c ccccc}
\toprule
& & & \multicolumn{5}{c}{\textbf{\dataset}} \\
\cmidrule{4-8}
\textbf{Uncond. Portion} & \textbf{Sketch Cond.} & & 
\textbf{FID} $\downarrow$ & \textbf{KID} $\downarrow$ & \textbf{CLIP-S} $\uparrow$  & \textbf{PD} $\downarrow$ & \textbf{SD} $\downarrow$ \\
\midrule
0.1 & 1.0 & & 9.64 & 3.76 & 30.24 & 7.66 & 0.459 \\
0.2 & 1.0 & & 8.62 & 3.24 & 29.06 & 7.51 & 0.430 \\
0.3 & 1.0 & & 10.93 & 4.78 & 28.47 & 7.69 & 0.432 \\
\midrule
0.2 & 0.8 & & 8.56 & 3.28 & 29.31 & 7.32 & 0.433 \\
0.2 & 0.6 & & 8.43 & 3.21 & 29.51 & 7.32 & 0.436 \\
0.2 & 0.4 & & 8.11 & 3.00 & 29.79 & 7.13 & 0.440 \\
\rowcolor{blond}
0.2 & 0.2 & & 7.73 & 2.82 & 30.04 & 6.79 & 0.458 \\
0.2 & 0.0 & & 7.82 & 2.85 & 29.93 & 6.26 & 0.519 \\
\bottomrule
\end{tabular}
}
\end{center}
\vspace{-0.4cm}
\caption{Ablation analysis of our complete model varying the unconditioning portion during training and the sketch conditioning steps. Results refer to the unpaired setting.
}
\vspace{-0.35cm}
\label{tab:dressCode_ablations}
\end{table}

\tit{Unconditional Training and Sketch Conditioning} In Table~\ref{tab:dressCode_ablations}, we inquire about the fully conditioned network performance according to the variance of the portion of unconditional training. Additionally, we evaluate the results by varying the fraction of sketch conditioning steps. As can be seen, the best results are achieved by using 0.2 for both parameters. In particular, for unconditional training, we train three different models (\ie~with 0.1, 0.2, and 0.3). When evaluating the sketch conditioning parameter, we test our model with values between 0 and 1 with a stride of 0.2. It is worth noting that the sketch distance consistently decreases as the number of sketch conditioning steps increases, showing the robustness of the approach.

\section{Conclusion}
\label{sec:conclusion}
The Multimodal Garment Designer proposed in this paper is the first latent diffusion model defined for human-centric fashion image editing, conditioned by multimodal inputs such as text, body pose, and sketches. 
The novel architecture, trained on two new semi-automatically annotated datasets and evaluated with standard and newly proposed metrics, as well as by user studies, is very promising. The result is one of the first successful attempts to mimic the designers' job in the creative process of fashion design and could be a starting point for a capillary adoption of diffusion models in creative industries, oversight by human input.

\section*{Acknowledgments}
This work has partially been supported by the European Commission under the PNRR-M4C2 project ``FAIR - Future Artificial Intelligence Research'' and the European Horizon 2020 Programme (grant number 101004545 - ReInHerit), and by the PRIN project ``CREATIVE: CRoss-modal understanding and gEnerATIon of Visual and tExtual content'' (CUP B87G22000460001), co-funded by the Italian Ministry of University.

{\small
\bibliographystyle{ieee_fullname}
\bibliography{bibliography}
}

\appendix
\section{\dataset and \datasetviton Datasets}
In this section, we give additional details about the dataset collection and annotation process and provide statistics and further examples of the collected datasets.

\subsection{Data Preparation}
\begin{figure}[t]
\centering
\includegraphics[width=0.98\linewidth]{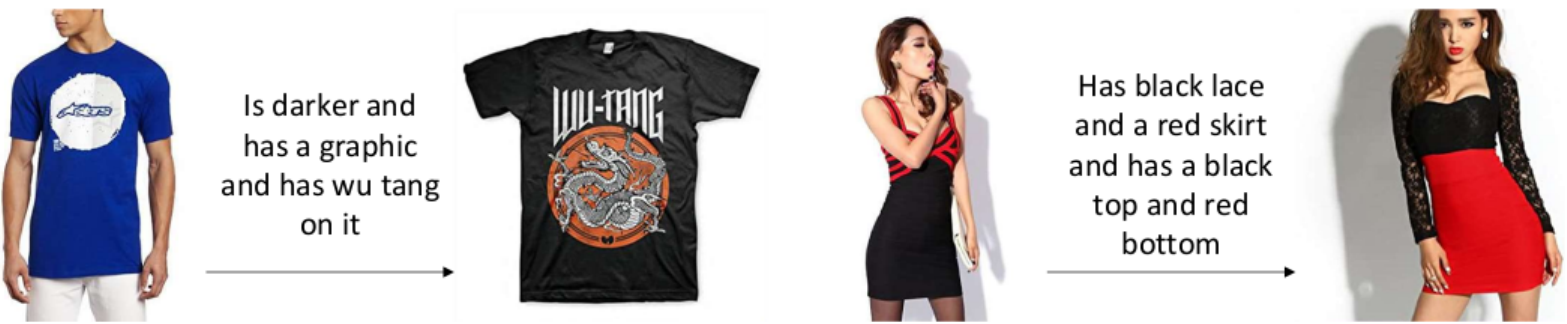}
\vspace{-0.25cm}
\caption{Examples of FashionIQ data type.}
\label{fig:fashionIQ_example}
\vspace{-0.2cm}
\end{figure}

Before extracting noun chunks from the textual sentences of FashionIQ~\cite{wu2021fashion} and Fashion200k~\cite{han2017automatic}, we perform word lemmatization to reduce each word to its root form. 
Such pre-processing stage is crucial for the FashionIQ dataset, as the captions do not describe a single garment but instead express the properties to modify in a given image to match its target. Fig.~\ref{fig:fashionIQ_example} shows two examples of FashionIQ annotations.

We use the spaCy NLP toolkit\footnote{\href{https://spacy.io/}{https://spacy.io/}} to extract noun chunks from textual sentences. To facilitate prompt engineering at a later stage, we remove the articles at the beginning of each noun chunk. 
Subsequently, we filter out all noun chunks starting with or containing special characters and keep unique elements.
Table~\ref{tab:iqand200_numbers} reports detailed statistics about the number of unique captions and extracted noun chunks from which we start the annotation.

\tit{Textual Prompts}
As described in the main paper, we rely on the cosine similarity between CLIP-based image and text embeddings to associate each garment with the 25 most representative noun chunks. We exploit prompt ensembling to perform such zero-shot association as it is shown in~\cite{Radford2021LearningTV} that this technique improves performance.

The employed textual prompts are: 
\begin{itemize}[noitemsep,topsep=0pt]
\item \texttt{\footnotesize a photo of a [noun chunk]},
\item \texttt{\footnotesize a photo of a nice [noun chunk]},
\item \texttt{\footnotesize a photo of a cool [noun chunk]},
\item \texttt{\footnotesize a photo of an expensive [noun chunk]},
\item \texttt{\footnotesize a good photo of a [noun chunk]},
\item \texttt{\footnotesize a bright photo of a [noun chunk]},
\item \texttt{\footnotesize a fashion studio shot of a [noun chunk]},
\item \texttt{\footnotesize a fashion magazine photo of a [noun chunk]},
\item \texttt{\footnotesize a fashion brochure photo of a [noun chunk]},
\item \texttt{\footnotesize a fashion catalog photo of a [noun chunk]},
\item \texttt{\footnotesize a fashion press photo of a [noun chunk]},
\item \texttt{\footnotesize a yoox photo of a [noun chunk]},
\item \texttt{\footnotesize a yoox web image of a [noun chunk]},
\item \texttt{\footnotesize a high-resolution photo of a [noun chunk]},
\item \texttt{\footnotesize a cropped photo of a [noun chunk]},
\item \texttt{\footnotesize a close-up photo of a [noun chunk]}, 
\item \texttt{\footnotesize a photo of one [noun chunk]}.
\end{itemize}

\begin{table}[t]
\begin{center}
\footnotesize
\setlength{\tabcolsep}{.32em}
\resizebox{\linewidth}{!}{
\begin{tabular}{lc ccc c ccc}
\toprule
& & \multicolumn{3}{c}{\textbf{Unique Captions}} & & \multicolumn{3}{c}{\textbf{Unique Noun Chunks}} \\
\cmidrule{3-5} \cmidrule{7-9}
\textbf{Dataset} & & \textbf{Upper} & \textbf{Lower} & \textbf{Dresses} & & \textbf{Upper} & \textbf{Lower} & \textbf{Dresses} \\
\midrule
FashionIQ~\cite{wu2021fashion} & & 27,339 & 0 & 15,101 & & 7,801 & 0 & 3,592  \\
Fashion200k~\cite{han2017automatic} & & 25,959 & 11,022 & 16,694 & & 22,898 & 13,420 & 15,890 \\
\bottomrule
\end{tabular}
}
\end{center}
\vspace{-0.4cm}
\caption{Number of unique captions and noun chunks for each category of the FashionIQ and Fashion200k datasets.}
\label{tab:iqand200_numbers}
\vspace{-0.45cm}
\end{table}

\begin{figure*}[t]
    \centering
    \begin{subfigure}{0.485\linewidth}
        \includegraphics[width=\linewidth]{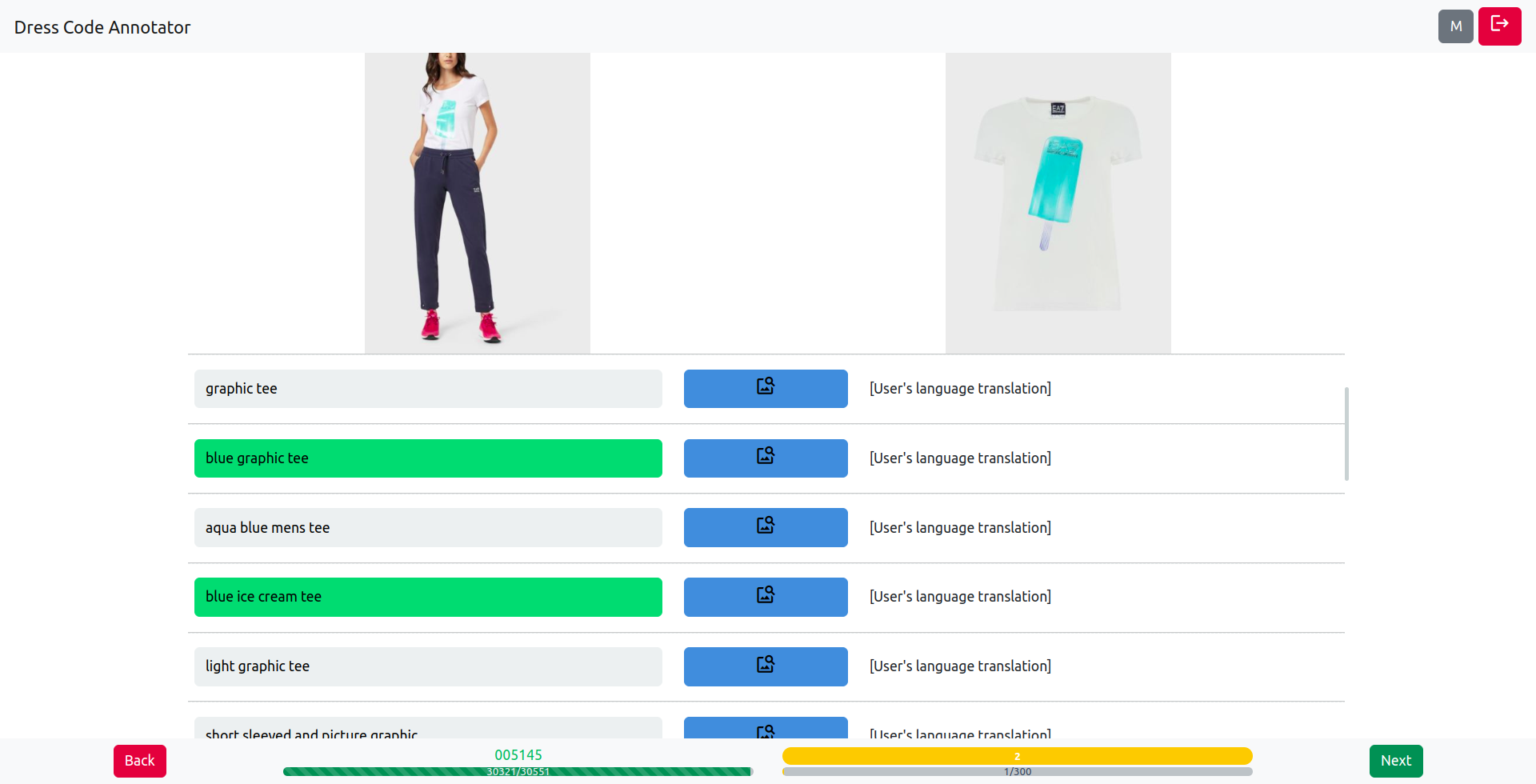}
        \caption{}
        \label{fig:caption_choice}
    \end{subfigure}
    \hspace{0.1cm}
    \begin{subfigure}{0.485\linewidth}
        \includegraphics[width=\linewidth]{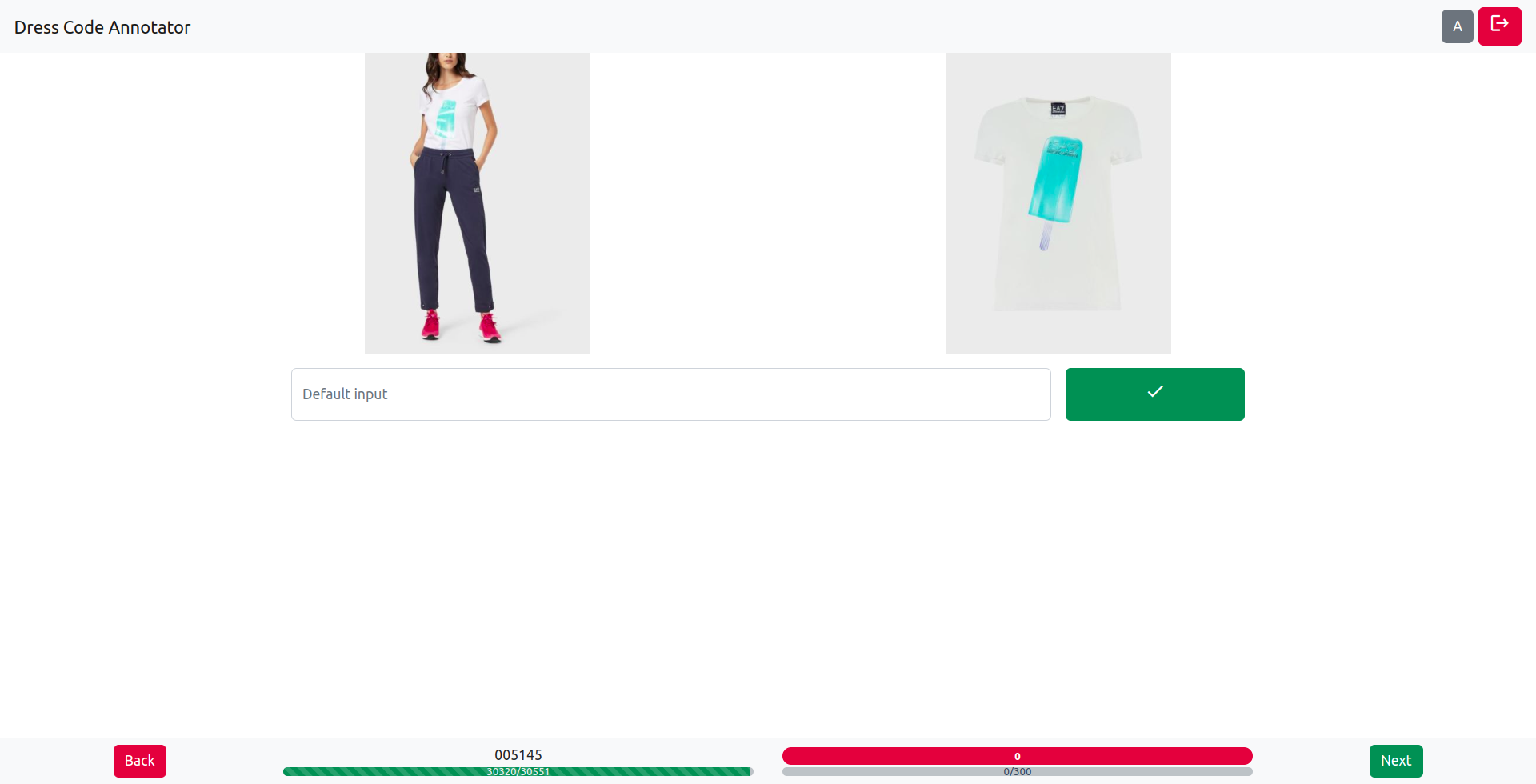}
        \caption{}
        \label{fig:custom_caption}
    \end{subfigure}
    \vspace{-0.35cm}
    \caption{User interface of the custom annotation tool. In (a) the user can select the noun chunks among the proposed ones, while in (b) the user can manually annotate the garment.}
    \label{fig:annotation_tool}
\vspace{-0.4cm}
\end{figure*}

\blfootnote{$^*$Equal contribution.}
\subsection{Annotation Tool for Fine-Grained Annotation}
We develop a custom annotation tool using the Django and Angular web frameworks to ease and speed up the fine-grained annotation process. 
Fig.~\ref{fig:annotation_tool} depicts the user interface.
In the annotation phase, users are provided with both model's image and the corresponding in-shop garment and should select the three most representative noun chunks per item (Fig.~\ref{fig:caption_choice}). 
If the automatic selection process fails to suggest three correct noun chunks, the user can manually insert them (Fig.~\ref{fig:custom_caption}).

\subsection{Coarse-Grained Annotation}
After completing the manual annotation process on Dress Code, we obtain 26,400 different model-garment pairs (with 8,800 items per category), each associated with three different noun chunks.
To annotate the remaining 27,392 items of \dataset and the 13,679 items of \datasetviton, we leverage the manually annotated image-text pairs and finetune the OpenCLIP ViT-B/32~\cite{wortsman2022robust} model pre-trained on the English portion of the LAION-5B dataset.

\tit{CLIP Finetuning}
We finetune both encoders of the OpenCLIP model using a single NVIDIA A100 GPU for 400 steps, with a batch size of 2048 and a learning rate of $10^{-6}$. As optimizer, we use AdamW~\cite{loshchilov2019decoupled} with a weight decay of 0.2.
We use mixed precision~\cite{micikevicius2018mixed} to speed up training and save memory.
During the training process, we monitor the model performance using the top-3 accuracy metric on the test split of the \dataset dataset. 
We choose this metric intending to associate each image with three distinct noun chunks. The out-of-the-box model achieves a top-3 accuracy of 12.95\%, which improves to 16.60\% after finetuning. The OpenCLIP ViT-g/14 model instead achieves a top-3 accuracy of 16.21\%, while being computationally heavier than the ViT-B/32 version. Since the ViT-g/14 model predicts the set of noun chunks from which we extract the ground-truth, the actual difference in performance between the finetuned ViT-B/32 model and the out-of-the-box ViT-g/14 model could be even higher.

\subsection{Extracting Sketches}
As mentioned in the main paper, we train a warping module to generate input sketches for the unpaired setting (\ie~when we give as input the multimodal information corresponding to a garment different from the one originally worn by the model). In particular, our method involves the transformation of a given in-shop garment $C \in \mathbb{R}^{H \times W \times 3}$ into a warped image of the same garment that fits the model of a target image $I$. We employ the warping module proposed in~\cite{wang2018toward}, refining the results with a U-Net based component~\cite{ronneberger2015u}.

The warping module computes a correlation map between the encoded representations of the in-shop garment $C$ and a cloth-agnostic person representation composed of the pose map $P \in \mathbb{R}^{H \times W \times 18}$ and the masked model image $I_M \in \mathbb{R}^{H \times W \times 3}$. We use two separate convolutional networks to obtain these encoded representations. Based on the computed correlation map, we predict the spatial transformation parameters $\theta$ of a thin-plate spline geometric transformation~\cite{rocco2017convolutional} (\ie~$\text{TPS}_\theta$).
We then use the $\theta$ parameters to compute the coarse warped garment $\hat{C}$ starting from the in-shop garment $C$ as follows:
\begin{equation}
    \hat{C} = \text{TPS}_\theta(C).
    \label{eq:tps_warping}
\end{equation}
To refine the result, we employ a U-Net model that takes as input the concatenation of the coarse warped garment $\hat{C}$, the pose map $P$, and the masked model image $I_M$, and predicts the refined warped garment $\Tilde{C}$.

\begin{table}[t]
\begin{center}
\footnotesize
\setlength{\tabcolsep}{.32em}
\resizebox{\linewidth}{!}{
\begin{tabular}{lccc ccc c ccc}
\toprule
& & & & \multicolumn{3}{c}{\textbf{Images}} & & \multicolumn{3}{c}{\textbf{Unique Noun Chunks}} \\
\cmidrule{5-7} \cmidrule{9-11}
\textbf{Dataset} & \textbf{Ann.} & \textbf{Split} & & \textbf{Upper} & \textbf{Lower} & \textbf{Dresses} & & \textbf{Upper} & \textbf{Lower} & \textbf{Dresses} \\
\midrule
\multirow{4}{*}{Dress Code M.} & \multirow{4}{*}{F} 
 & Train & & 7,000 & 7,000 & 7,000 & & 4,751 & 5,914 & 4,410  \\
 & & Test & & 1,800 & 1,800 & 1,800 & & 2,337 & 2,861 & 2,144  \\
 & & $ \cup $ & & 8,800 & 8,800 & 8,800 & & 5,284 & 6,509 & 4,915  \\
 & & $ \cap $ & & - & - & - & & 1,804 & 2,266 & 1,639  \\
 \midrule
\multirow{4}{*}{Dress Code M.} & \multirow{4}{*}{C} 
 & Train & & 6,563 & 151 & 20,666 & & 7,198 & 320 & 8,650  \\
 & & Test & & 0 & 0 & 0 & & 0 & 0 & 0  \\
 & & $ \cup $ & & 6,563 & 151 & 20,666 & & 7,198 & 320 & 8,650  \\
 & & $ \cap $ & & - & - & - & & 0 & 0 & 0  \\
 \midrule
 \multirow{4}{*}{Dress Code M.} & \multirow{4}{*}{F+C}
 & Train & & 13,563 & 7,151 & 27,666 & & 9,163 & 6,037 & 9,465  \\
 & & Test & & 1,800 & 1,800 & 1,800 & & 2,337 & 2,861 & 2,144  \\
 & & $ \cup $ & & 15,363 & 8,951 & 29,466 & & 9431 & 6,597 & 9,568  \\
 & & $ \cap $ & & - & - & - & & 2,069 & 2,301 & 2,041  \\
 \midrule
 \multirow{4}{*}{VITON-HD M.} & \multirow{4}{*}{C}
 & Train & & 11,647 & - & - & & 4,823 & - & -  \\
 & & Test & & 2,032 & - & - & & 2,149  & - & -  \\
 & & $ \cup $ &  & 13,679 & - & - & & 5,143 & - &  - \\
 & & $ \cap $ & & - & - & - & & 1,829  & - & - \\
\bottomrule
\end{tabular}
}
\end{center}
\vspace{-0.4cm}
\caption{Number of images and unique noun chunks per category for both \dataset and \datasetviton. (F) indicates the fine-grained annotation while (C) indicates the coarse-grained annotation.}
\label{tab:fineandcoarse_numbers}
\vspace{-0.2cm}
\end{table}

\begin{figure}[t]
\begin{center}
\includegraphics[width=\linewidth]{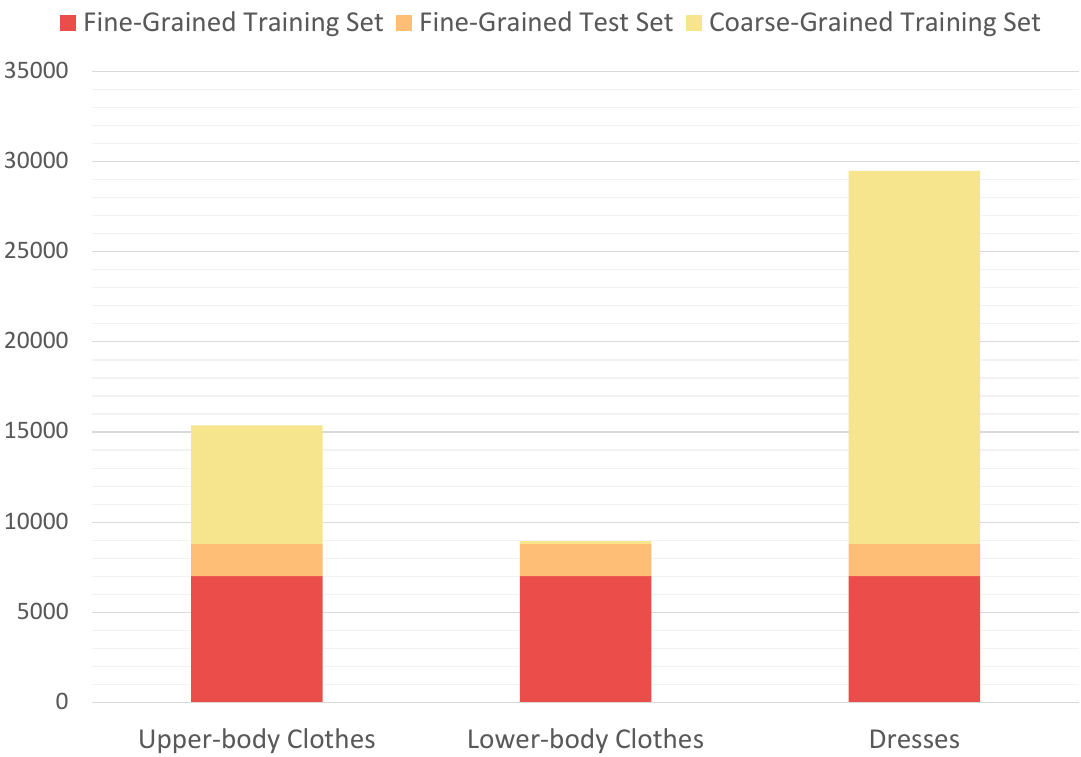}
\end{center}
\vspace{-0.4cm}
\caption{Annotated images per category on Dress Code Multimodal.}
\label{fig:dresscode_dataset_statistics}
\vspace{-0.3cm}
\end{figure}

We train this model on the training set of both \dataset and \datasetviton using a combination of an L1 loss between generated and target in-shop garments and a perceptual loss (also known as VGG loss~\cite{johnson2016perceptual}) to compute the difference between the feature maps of generated and target garments extracted with a VGG-19~\cite{simonyan2014very}. We train with a resolution of $256 \times 192$, Adam~\cite{kingma2015adam} as optimizer with $\beta_1 = 0.5, \beta_2 = 0.99$, and a learning rate equal to $10^{-4}$. We train the network on the VITON-HD dataset for 30 epochs, while the training on the Dress Code dataset converges after 80 epochs.

\subsection{Additional Statistics and Annotated Samples}
Table~\ref{tab:fineandcoarse_numbers} summarizes the number of images and unique noun chunks for each category of \dataset and \datasetviton.
The table shows that the datasets share noun chunks between the train and test set ($ \cap $). This behavior is likely due to the limited capacity of the textual modality to represent the whole semantic information of the image. Fig.~\ref{fig:dresscode_dataset_statistics} instead shows the number of samples for each category highlighting the different annotation strategies on \dataset. 

\begin{figure}[t]
    \centering
    \hspace*{\fill}
    \begin{subfigure}[b]{0.3\linewidth}
        \includegraphics[width=\linewidth]{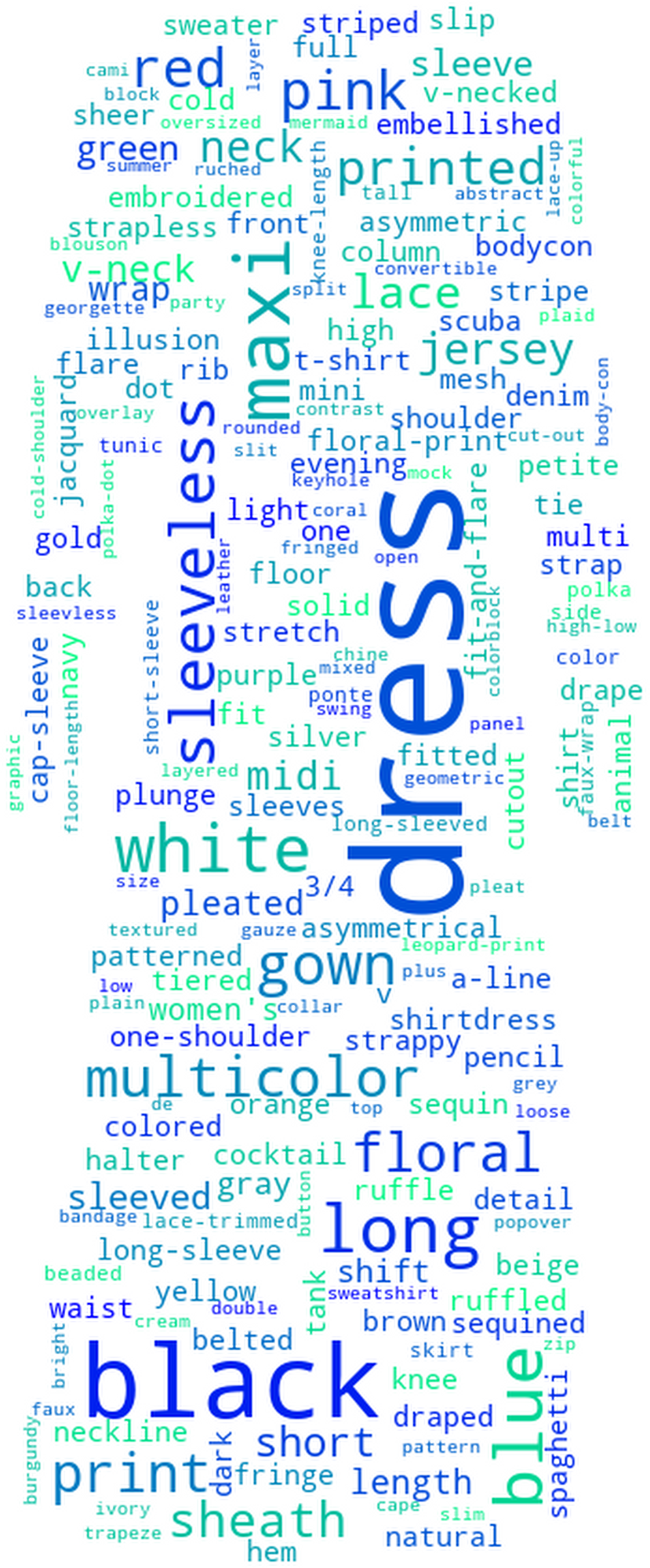}
        \caption{}
        \label{fig:wordcount_dresses}
    \end{subfigure}
    \hfill
    \begin{subfigure}[b]{0.3\linewidth}
        \includegraphics[width=\linewidth]{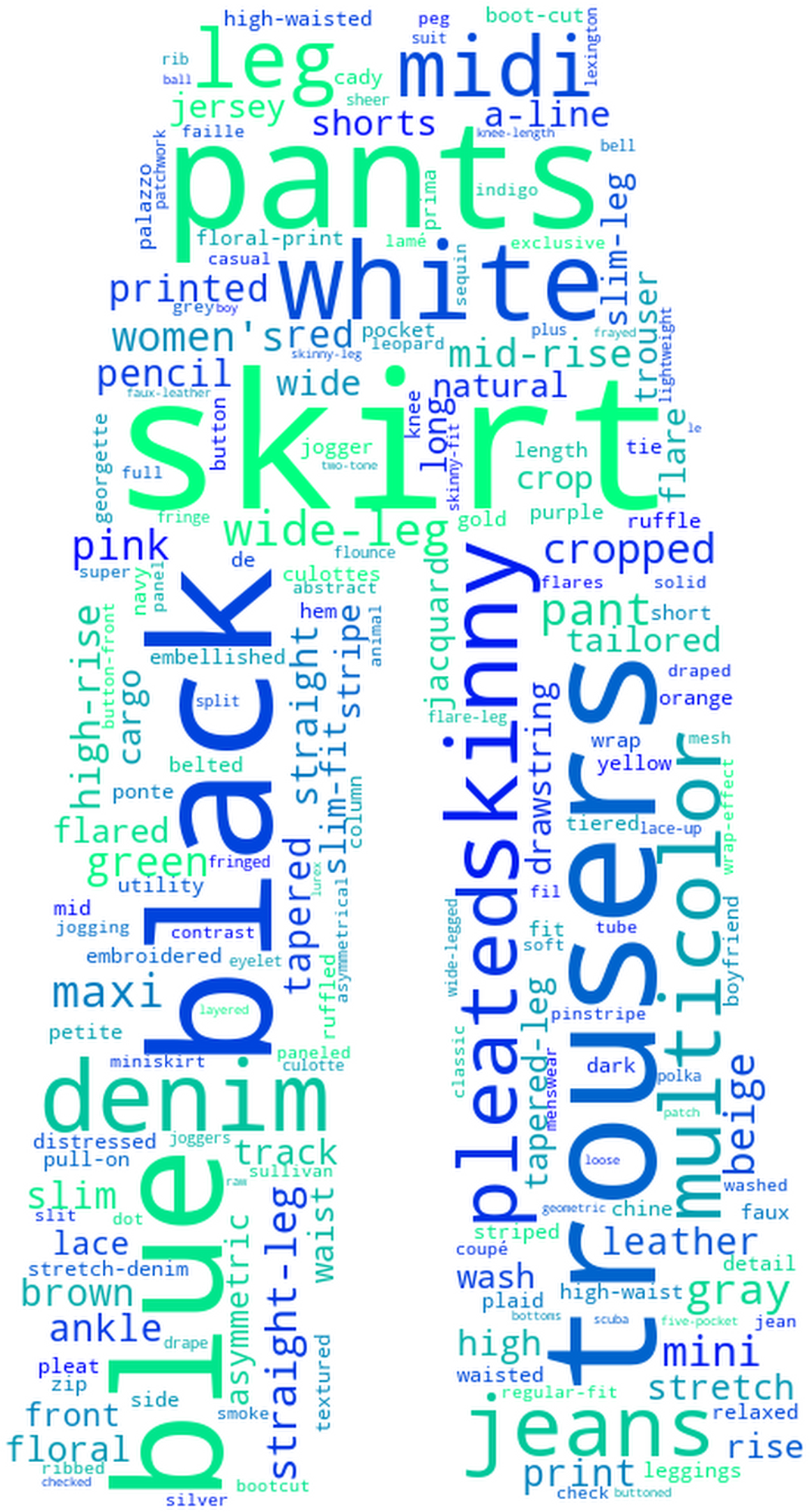}
        \caption{}
        \label{fig:wordcount_lower}
    \end{subfigure}
    \hspace*{\fill}
    
    \hspace*{\fill}
    \begin{subfigure}[b]{0.3\linewidth}
        \includegraphics[width=\linewidth]{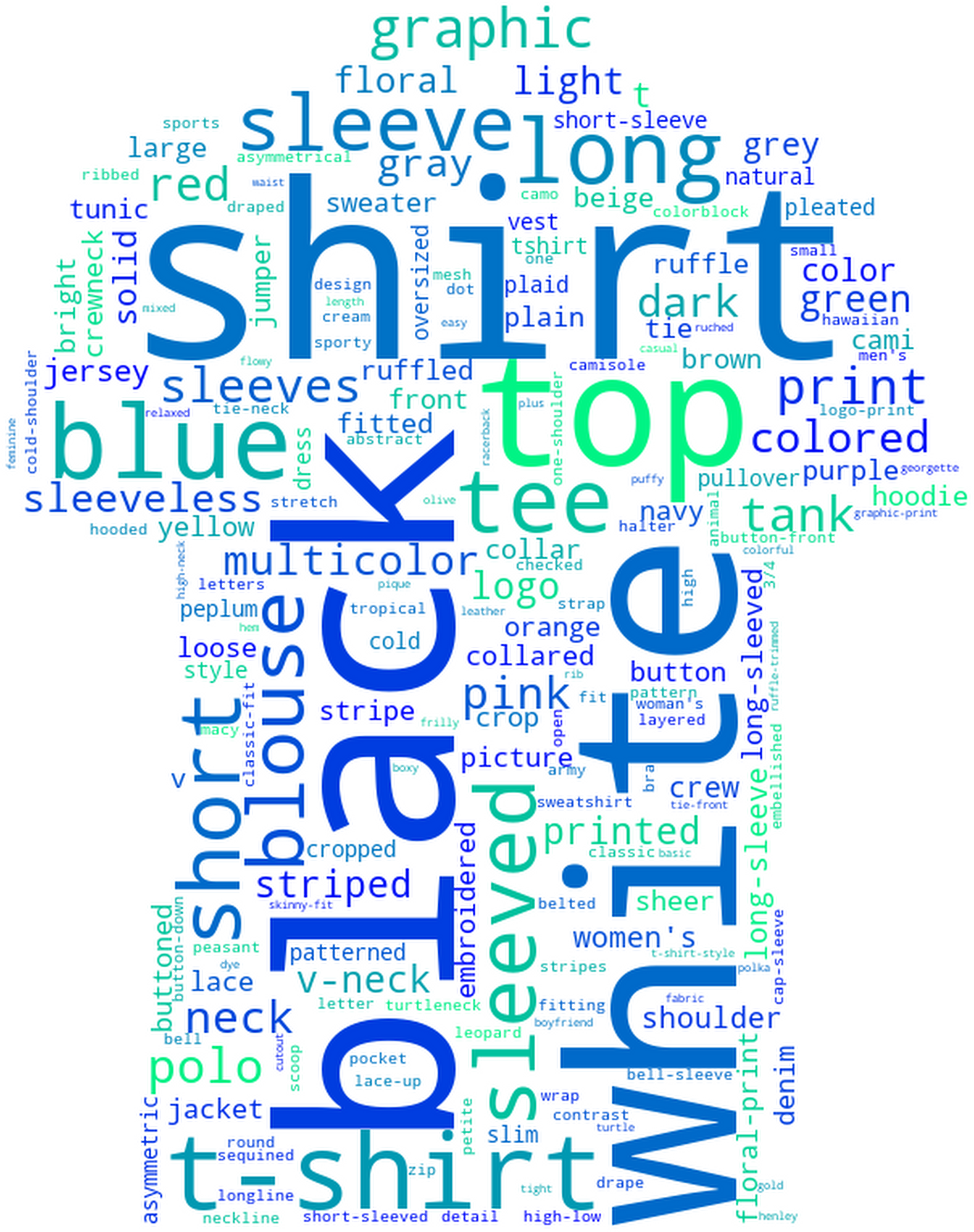}
        \caption{}
        \label{fig:wordcount_upper}
    \end{subfigure}
    \hfill
    \begin{subfigure}[b]{0.3\linewidth}
    \includegraphics[width=\linewidth]{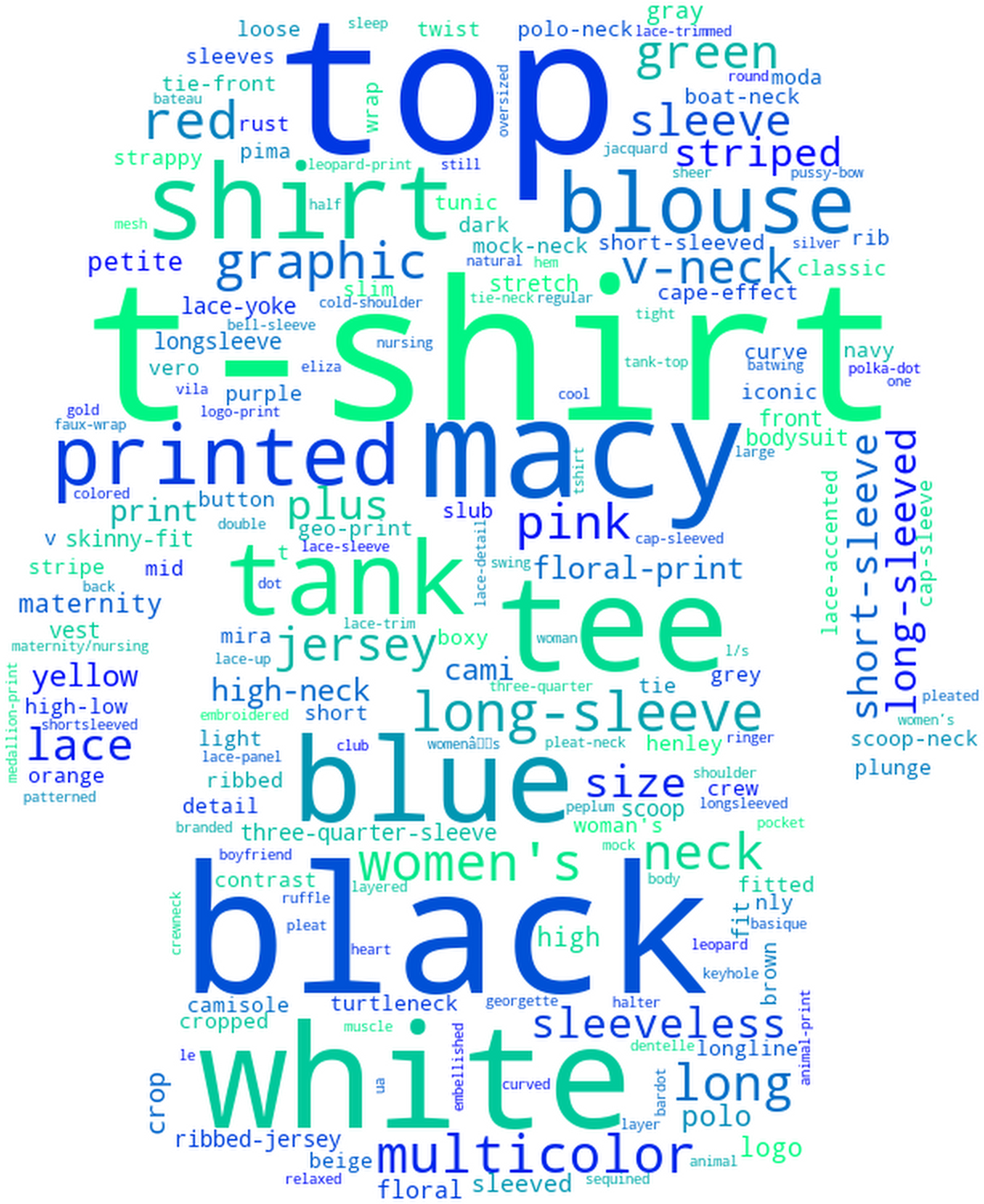}
    \caption{}
    \label{fig:wordcount_viton}
    \end{subfigure}
    \hspace*{\fill}
\vspace{-.3cm}
     \caption{Vocabulary of the frequent words scaled by frequency for dresses (a), lower-body clothes (b), upper-body clothes (c) of \dataset and clothing items of \datasetviton (d).}
     \label{fig:wordcount}
    \vspace{-.3cm}
\end{figure}

In Fig.~\ref{fig:wordcount}, we report the word clouds extracted from the textual annotations, representing the most frequently used words in the collected noun chunks for each category of \dataset and \datasetviton. From this visualization, we can notice that the frequency of the terms varies according to the garment category, and the semantic richness of our annotations is consistent across different garment types. 

\begin{figure*}
\begin{center}
\resizebox{0.98\linewidth}{!}{
\includegraphics[width=\linewidth]{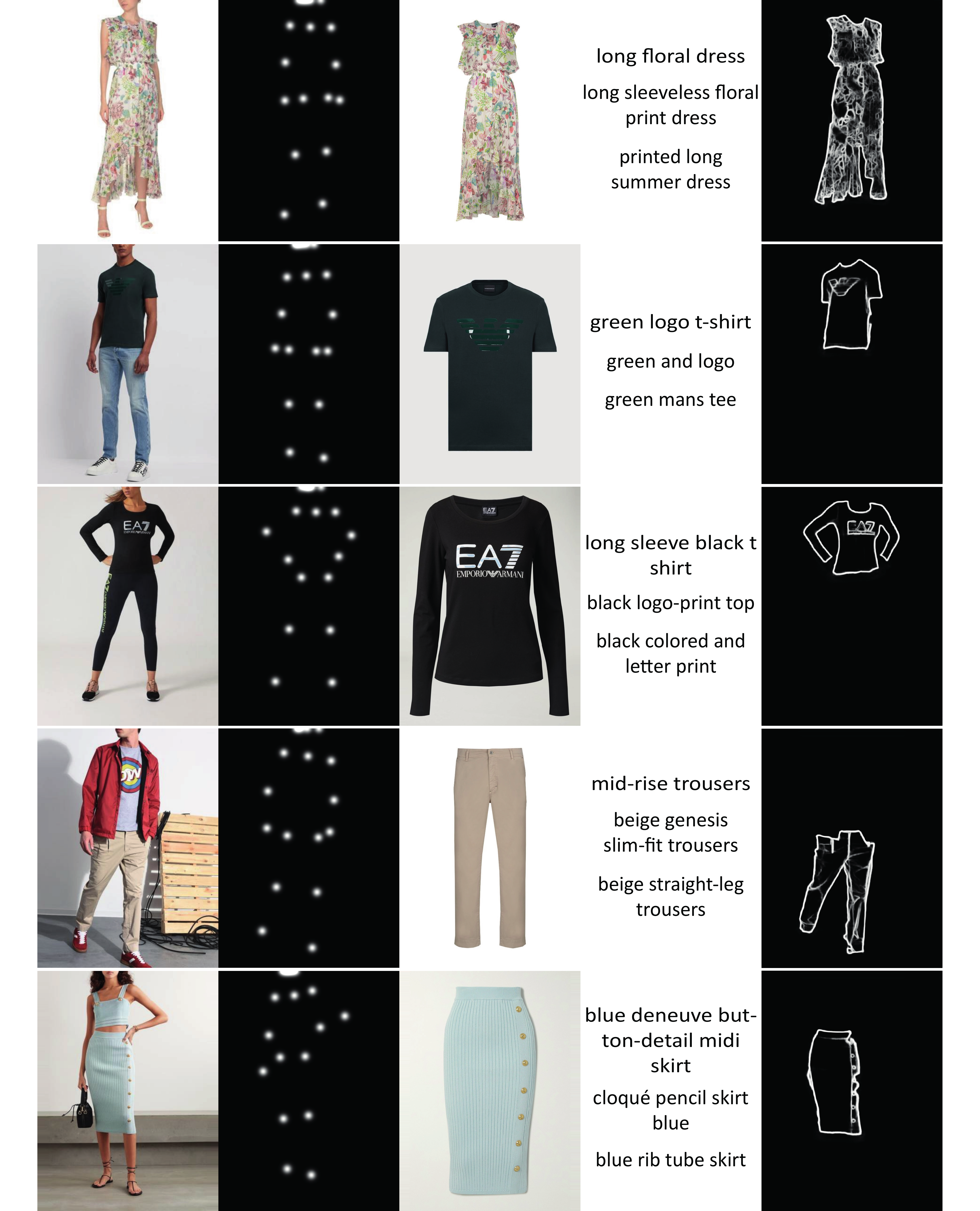}
}
\end{center}
\vspace{-0.4cm}
\caption{Sample images and multimodal data from our newly collected \dataset dataset (fine-grained textual annotations).}
\label{fig:dresscode_fine_samples}
\end{figure*}

\begin{figure*}
\begin{center}
\resizebox{0.98\linewidth}{!}{
\includegraphics[width=\linewidth]{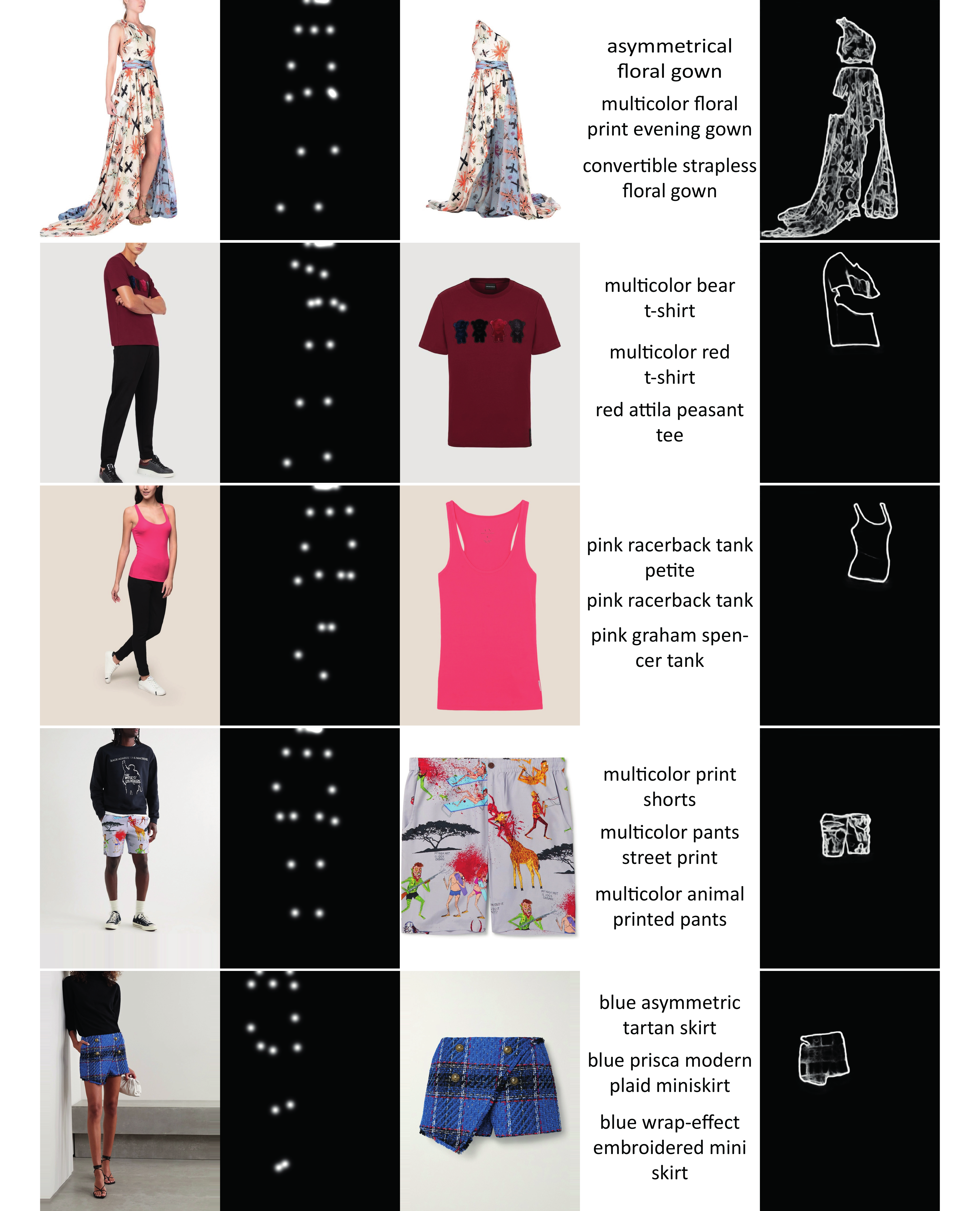}
}
\end{center}
\vspace{-0.4cm}
\caption{Sample images and multimodal data from our newly collected \dataset dataset (coarse-grained textual annotations).}
\label{fig:dresscode_coarse_samples}
\end{figure*}

\begin{figure*}
\begin{center}
\resizebox{0.98\linewidth}{!}{
\includegraphics[width=\linewidth]{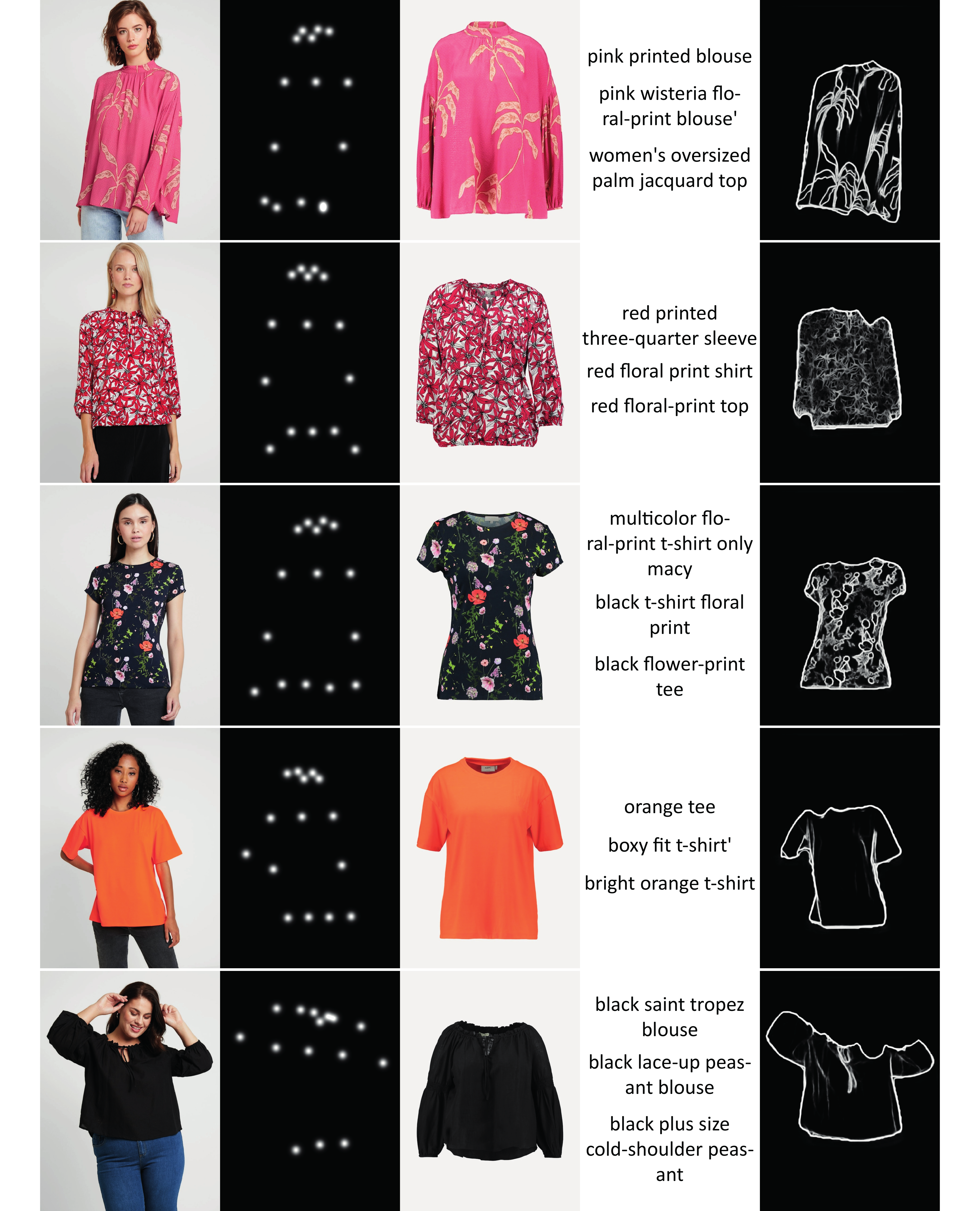}
}
\end{center}
\vspace{-0.4cm}
\caption{Sample images and multimodal data from our newly collected \datasetviton dataset (coarse-grained textual annotations).}
\label{fig:vitonhd_coarse_samples}
\end{figure*}

In Fig.~\ref{fig:dresscode_fine_samples} and Fig.~\ref{fig:dresscode_coarse_samples}, we report samples from the fine-grained and coarse-grained subsets of \dataset, respectively.
Instead, in Fig.~\ref{fig:vitonhd_coarse_samples}, we show additional examples extracted from \datasetviton.

\section{Evaluation Metrics}
This section provides additional details about the evaluation metrics used in our experiments. We first give some clarifications about the CLIP-S metric and then present an accurate formulation of the proposed sketch distance and pose distance metrics.

\tit{CLIP-S}
The CLIP score~\cite{hessel2021clipscore} is a well-known metric to evaluate the similarity between images and textual sentences. In our setting, we employ this metric to assess the coherence of the generated images with respect to the corresponding textual inputs used to condition the generation process. As mentioned in the main paper, our implementation relies on the CLIP-S of the TorchMetrics library~\cite{detlefsen2022torchmetrics} and adopts the ViT-H/14 trained on LAION-2B as the CLIP model. Specifically, we crop the generated image using the bounding box used to mask it and paste the resulting crop on a white background, obtaining a final resolution equal to $224\times224$.
The adopted metric is defined as follows:
\begin{equation}
\text{CLIP-S}(I,Y) = \max (100*cos(E_{\Tilde{I}},E_Y),0),
\label{eq:CLIP_score}
\end{equation}
where $E_{\Tilde{I}}$ represents the CLIP embedding of the generated portion of the image $\Tilde{I}$ pasted on white background, $E_Y$ represents the CLIP embedding for the caption $Y$, and $cos$ is the cosine similarity. We calculate the cosine similarity between the image and caption embeddings and scale the result by a factor of 100. If the cosine similarity is negative, then CLIP-S is zero. 

\tit{Pose Distance (PD)} 
To measure the coherence of human-body poses between the generated image and the original one, we propose a novel pose distance metric that estimates the distance between human keypoints extracted from the original and the generated images. 
Given a ground-truth image $I$ and a generated image $\Tilde{I}$, we extract human keypoints from each of them using the keypoint extraction network $\mathcal{K}$ (\ie~in our case, we use OpenPifPaf~\cite{kreiss2021openpifpaf}) and identify the set of keypoints falling in the mask $M$ as $\mathcal{K}(\cdot)_M$. We compute the final score with an $\ell_2$ distance between each pair of real-generated corresponding keypoints (\ie~$k \in \mathcal{K}(I)_M$ and $\Tilde{k} \in \mathcal{K}(\Tilde{I})_M$, respectively), weighting each keypoint distance with the detector confidence to consider possible estimation errors. Formally, our pose distance metric is defined as follows:
\begin{equation}
\text{PD}(I, \Tilde{I}) = \frac{\sum\limits_{\substack{k \in \mathcal{K}(I)_M \\ \Tilde{k} \in \mathcal{K}(\Tilde{I})_M}} \sqrt{(k_x-\Tilde{k}_x)^2 + (k_y-\Tilde{k}_y)^2}  \cdot \text{CF}_{k\Tilde{k}}}{\sum_{k\Tilde{k}} \text{CF}_{k\Tilde{k}}},
\end{equation}
where, for each pair of real-generated keypoints, $\text{CF}_{k\Tilde{k}}$ is $1$ if the confidence of the detector $\mathcal{K}$ on both keypoints is greater or equal to $0.5$, and $0$ otherwise.
 
\tit{Sketch Distance (SD)}
To evaluate the adherence of the generated images to the constraints imposed by the input sketch, we propose a new sketch distance metric. To compute the metric, we first extract the ground-truth and the generated garments label maps using an off-the-shelf semantic segmentation model\footnote{\href{https://github.com/levindabhi/cloth-segmentation}{https://github.com/levindabhi/cloth-segmentation}}. We segment the garment according to its category and paste it on a white background of shape $512\times384$. We refer to these new images with $I_S$ and $\Tilde{I}_S$, respectively. Then, we extract the garment sketches of both the ground-truth and the generated images using an edge detector network $Edge$ (\ie~PIDInet~\cite{su2021pixel}). Finally, we compute the mean squared error between the extracted sketches, weighting the per-pixel results on the inverse frequency of the activated pixels. Formally, the introduced sketch distance metric is defined as follows:
\begin{equation}
\text{SD}(I_S, \Tilde{I}_S) = \text{MSE}\left(Edge(I_S), Edge(\Tilde{I}_S)\right) * p,
\label{eq:sketch_metric}
\end{equation}
where $p$ is the inverse pixel frequency. It is noteworthy that sketch thresholding could be applied before distance computation. Nevertheless, we argue that avoiding thresholding enables an effective comparison of hand-drawn ground-truth grayscale sketches. This approach can facilitate the evaluation of methods that generate images conditioned using the sketch. Therefore, we think the proposed metric can be a valuable tool for comparing sketch-guided generative architectures.

\begin{figure}[t]
\centering
\begin{subfigure}{\linewidth}\centering
    \includegraphics[width=0.8\linewidth]{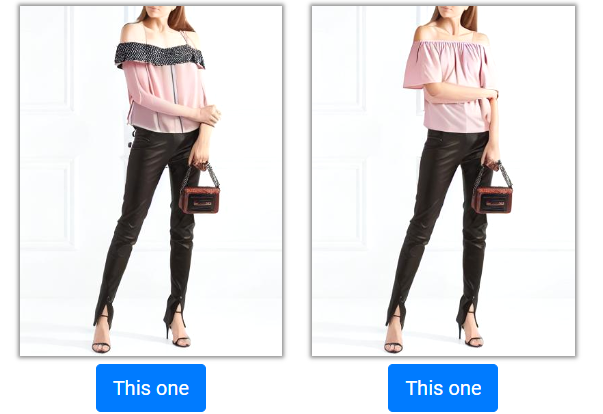}
    \caption{}
    \label{fig:realism_test}
\end{subfigure}
\begin{subfigure}{\linewidth}
    \includegraphics[width=\linewidth]{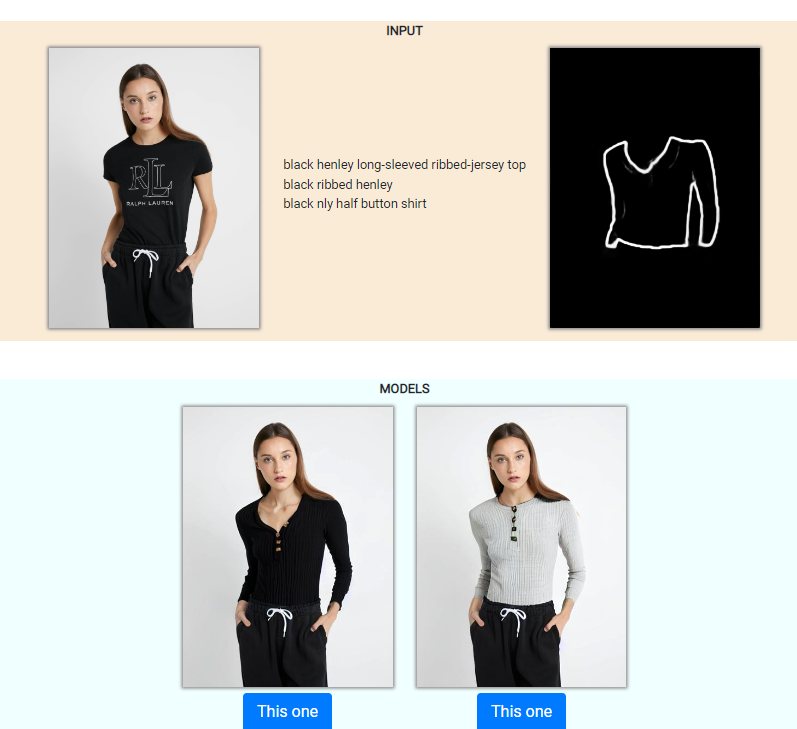}
    \caption{}
    \label{fig:adherence_test}
\end{subfigure}
\vspace{-0.5cm}
\caption{User study interface, where (a) corresponds to the realism evaluation and (b) refers to the coherence analysis between generated images and the given multimodal inputs.}
\label{fig:user_study}
\vspace{-0.3cm}
\end{figure}

\begin{table*}[ht!]
\begin{center}
\footnotesize
\setlength{\tabcolsep}{.32em}
\resizebox{\linewidth}{!}{
\begin{tabular}{lcc ccc c ccccc c ccccc c ccccc}
\toprule
 & & & \multicolumn{3}{c}{\textbf{Modalities}} & & \multicolumn{5}{c}{\textbf{Upper-body}} & & \multicolumn{5}{c}{\textbf{Lower-body}} & & \multicolumn{5}{c}{\textbf{Dresses}} \\
\cmidrule{4-6} \cmidrule{8-12} \cmidrule{14-18} \cmidrule{20-24}
\textbf{Model} & \textbf{Resolution} & & \textbf{Text} & \textbf{Keypoints} & \textbf{Sketch} & & \textbf{FID} $\downarrow$ & \textbf{KID}  $\downarrow$ & \textbf{CLIP-S} $\uparrow$ & \textbf{PD} $\downarrow$ & \textbf{SD} $\downarrow$ & & \textbf{FID} $\downarrow$ & \textbf{KID} $\downarrow$ & \textbf{CLIP-S} $\uparrow$ & \textbf{PD} $\downarrow$ & \textbf{SD} $\downarrow$ & & \textbf{FID} $\downarrow$ & \textbf{KID} $\downarrow$ & \textbf{CLIP-S} $\uparrow$ & \textbf{PD} $\downarrow$ & \textbf{SD} $\downarrow$ \\
\midrule
\textit{Paired setting} \\
\hspace{0.4cm}Stable Diff.~\cite{rombach2022high} & 256$\times$192 & & \cmark & & & & 22.86 & 9.73 & 28.31 & 4.29 & - &  & 28.78 & 13.93 & 26.41 & 4.97 & - &  & 36.31 & 20.74 & 27.84 & 5.67 & -\\
\hspace{0.4cm}FICE~\cite{pernuvs2023fice} & 256$\times$192 & & \cmark & \cmark & & & 46.41 & 32.26 & 28.58 & 7.46 & - &  & 41.68 & 27.22 & 28.14 & 7.54 & - &  & 34.06 & 20.58 & 29.47 & 6.06 & - \\
\rowcolor{blond}
\hspace{0.4cm}\textbf{\ours (ours)} & 256$\times$192 & & \cmark & \cmark & & & \bf11.88 & \bf2.82 & \bf31.48 & \bf1.91 & - &  & \bf10.24 & \bf1.55 & \bf30.50 & \bf2.58 & - &  & \bf11.87 & \bf2.03 & \bf32.05 & \bf2.57 & - \\
\midrule
\textit{Paired setting} \\
\hspace{0.4cm}Stable Diff.~\cite{rombach2022high} & 512$\times$384 & & \cmark & & & & 21.00 & 8.59 & 30.17 & 7.95 & 0.310 &  & 28.40 & 14.48 & 28.02 & 9.96 & 0.345 &  & 33.12 & 17.39 & 29.36 & 9.86 & 0.450 \\
\hspace{0.4cm}SDEdit~\cite{meng2022sdedit} & 512$\times$384 & & \cmark & \cmark & \cmark & & 15.78 & 5.52 & 29.73 & 4.21 & 0.222 &  & 16.64 & 6.07 & 29.00 & 6.51 & 0.256 &  & 21.53 & 9.02 & 28.89 & 5.67 & 0.270 \\
\rowcolor{blond}
\hspace{0.4cm}\textbf{\ours (ours)} & 512$\times$384 & & \cmark & \cmark & \cmark & & \bf12.42 & \bf3.71 & \bf31.90 & \bf3.72 & \bf0.190 &  & \bf10.70 & \bf2.01 & \bf31.10 & \bf5.70 & \bf0.210 &  & \bf11.38 & \bf1.89 & \bf32.02 & \bf4.93 & \bf0.194 \\
\midrule
\textit{Unpaired setting} \\
\hspace{0.4cm}Stable Diff.~\cite{rombach2022high} & 256$\times$192 & & \cmark & & & & 22.86 & 9.73 & 28.31 & 4.29 & - &  & 28.78 & 13.93 & 26.41 & 4.97 & - &  & 36.31 & 20.74 & 27.84 & 5.67 & - \\
\hspace{0.4cm}FICE~\cite{pernuvs2023fice} & 256$\times$192 & & \cmark & \cmark & & & 49.77 & 35.37 & 26.48 & 7.64 & - &  & 44.94 & 30.39 & 25.42 & 7.84 & - &  & 39.04 & 25.27 & 26.14 & 6.39 & - \\
\rowcolor{blond}
\hspace{0.4cm}\textbf{\ours (ours)} & 256$\times$192 & & \cmark & \cmark & & & \bf14.50 & \bf3.48 & \bf29.24 & \bf2.39 & - &  & \bf13.70 & \bf2.48 & \bf29.09 & \bf3.32 & - &  & \bf13.72 & \bf2.50 & \bf30.37 & \bf3.17 & - \\
\midrule
\textit{Unpaired setting} \\
\hspace{0.4cm}Stable Diff.~\cite{rombach2022high} & 512$\times$384 & & \cmark & & & & 24.23 & 10.39 & 28.64 & 8.59 & 0.413 &  & 30.90 & 15.38 & 27.03 & 10.43 & 0.453 &  & 35.96 & 19.94 & 28.37 & 10.60 & 0.609 \\
\hspace{0.4cm}SDEdit~\cite{meng2022sdedit} & 512$\times$384 & & \cmark & \cmark & \cmark & & 17.86 & 6.50 & 27.36 & \bf4.78 & 0.357 &  & 19.16 & 6.85 & 27.08 & \bf7.53 & 0.399 &  & 22.97 & 9.98 & 26.85 & \bf6.42 & 0.411 \\
\rowcolor{blond}
\hspace{0.4cm}\textbf{\ours (ours)} & 512$\times$384 & & \cmark & \cmark & \cmark & & \bf15.99 & \bf4.50 & \bf29.76 & 5.41 & \bf0.291 &  & \bf14.82 & \bf2.81 & \bf29.96 & 7.96 & \bf0.289 &  & \bf14.71 & \bf3.63 & \bf30.41 & 7.15 & \bf0.252 \\
\bottomrule
\end{tabular}
}
\end{center}
\vspace{-0.4cm}
\caption{Category-wise quantitative results on the \dataset dataset.}
\label{tab:dresscode_categories}
\vspace{-0.35cm}
\end{table*}

\begin{table}[t]
\begin{center}
\footnotesize
\setlength{\tabcolsep}{.52em}
\resizebox{0.88\linewidth}{!}{
\begin{tabular}{c c ccccc}
\toprule
& & \multicolumn{5}{c}{\textbf{\dataset}} \\
\cmidrule{3-7}
\textbf{Sketch Cond.} & & 
\textbf{FID} $\downarrow$ & \textbf{KID} $\downarrow$ & \textbf{CLIP-S} $\uparrow$  & \textbf{PD} $\downarrow$ & \textbf{SD} $\downarrow$ \\
\midrule
1.0 & & 5.44 & 1.82 & 31.03 & 4.43 & 0.363 \\
0.8 & & 5.65 & 1.96 & 31.17 & 4.42 & 0.364 \\
0.6 & & 5.73 & 2.11 & 31.31 & 4.50 & 0.365 \\
0.4 &  & 5.80 & 2.17 & 31.44 & 4.51 & 0.368 \\
\rowcolor{blond}
0.2 &  & 5.74 & 2.11 & 31.68 & 4.72 & 0.374 \\
0.0 &  & 6.31 & 2.33 & 31.67 & 5.31 & 0.405 \\
\bottomrule
\end{tabular}
}
\end{center}
\vspace{-0.4cm}
\caption{Ablation study by varying the sketch conditioning steps on the paired setting of \dataset.}
\label{tab:dressCode_sketchCond}
\vspace{-0.2cm}
\end{table}

\section{User Study}
As mentioned in the main paper, we conduct a user study to evaluate the realism of generated images and their adherence to the given multimodal inputs, comparing our results with those from the considered competitors. To this aim, we develop a custom web interface presenting two different surveys. The former (Fig.~\ref{fig:realism_test}) assesses the realism of the generated output asking the user to select for each comparison the image that seems more realistic. In the latter (Figure~\ref{fig:adherence_test}), given the model's image, the set of noun chunks describing the garment, and the sketch, the user is asked to select which of the two proposed outputs looks more coherent with the multimodal inputs also taking into account the model's body pose. Overall, we collect around 7k evaluations, 3.5k for each test, and involving more than 150 users.

\begin{table}[t]
\begin{center}
\footnotesize
\setlength{\tabcolsep}{.52em}
\resizebox{0.88\linewidth}{!}{
\begin{tabular}{c c ccccc}
\toprule
& & \multicolumn{5}{c}{\textbf{\datasetviton}} \\
\cmidrule{3-7}
\textbf{Sketch Cond.} & & 
\textbf{FID} $\downarrow$ & \textbf{KID} $\downarrow$ & \textbf{CLIP-S} $\uparrow$  & \textbf{PD} $\downarrow$ & \textbf{SD} $\downarrow$ \\
\midrule
1.0 & & 13.01 & 4.00 & 30.32 & 7.05 & 0.225 \\
0.8 & & 12.75 & 3.73 & 30.46 & 7.11 & 0.250 \\
0.6 & & 12.76 & 3.75 & 30.53 & 7.13 & 0.263 \\
0.4 & & 12.71 & 3.67 & 30.56 & 7.12 & 0.280 \\
\rowcolor{blond}
0.2 & & 12.81 & 3.86 & 30.75 & 7.22 & 0.317 \\
0.0 & & 12.40 & 3.36 & 30.34 & 7.53 & 0.435 \\
\bottomrule
\end{tabular}
}
\end{center}
\vspace{-0.4cm}
\caption{Ablation study by varying the sketch conditioning steps on the unpaired setting of \datasetviton.}
\label{tab:vitonhd_sketchcond_unpaired}
\vspace{-0.35cm}
\end{table}
\section{Additional Results}
In this section, we provide additional experimental results to understand the strengths and limitations of our approach. Table~\ref{tab:dresscode_categories} extends Table~{\color{red}2} of the main paper showing quantitative results on each garment category of \dataset. Since each category contains only 1,800 images, the FID score presents a high variance in the results~\cite{binkowski2018demystifying}, while the KID metric presents more accurate results. Nevertheless, our method outperforms all competitors in all metrics except for the pose metrics in the unpaired setting. This behavior is due to the imperfect match of the predicted warped unpaired sketches and the model's body shape and pose. In fact, from the analysis of the sketch conditioning steps in the unpaired setting (Table~{\color{red}5} of the main paper), we can see that the pose distance directly correlates with the sketch conditioning parameter, while in the paired one (Table~\ref{tab:dressCode_sketchCond}) the pose distance metric decreases as the number of sketch conditioning steps increases. Instead, when evaluating the results on \datasetviton, the pose distance metric in the unpaired setting decreases (Table~\ref{tab:vitonhd_sketchcond_unpaired}). We believe this behavior relates to the size of the worn garment in this last dataset, which facilitates garment warping. In fact, VITON-HD features half-body images, while Dress Code contains full-body target models.

\begin{table}[t]
\begin{center}
\footnotesize
\setlength{\tabcolsep}{.3em}
\resizebox{\linewidth}{!}{
\begin{tabular}{lc ccc c ccccc}
\toprule
& & \multicolumn{3}{c}{\textbf{Modalities}} & & \multicolumn{5}{c}{\textbf{\dataset}} \\
\cmidrule{3-5} \cmidrule{7-11} 
\textbf{Model} & & \textbf{Text} & \textbf{Pose} & \textbf{Sketch} & & 
\textbf{FID} $\downarrow$ & \textbf{KID} $\downarrow$ & \textbf{CLIP-S} $\uparrow$  & \textbf{PD} $\downarrow$ & \textbf{SD} $\downarrow$ \\
\midrule
 & & \cmark & & & & \bf7.61 & \bf2.54 & \bf30.17 & 7.22 & 0.527 \\
 & & \cmark & \cmark & & & 7.82 & 2.85 & 29.93 & \bf6.26 & 0.519 \\
\rowcolor{blond}
\textbf{\ours (ours)} & & \cmark & \cmark & \cmark & & 7.73 & 2.82 & 30.04 & 6.79 & \bf0.458 \\
\midrule
& & \multicolumn{3}{c}{\textbf{Modalities}} & & \multicolumn{5}{c}{\textbf{\datasetviton}} \\
\cmidrule{3-5} \cmidrule{7-11} 
\textbf{Model} & & \textbf{Text} & \textbf{Pose} & \textbf{Sketch} & & 
\textbf{FID} $\downarrow$ & \textbf{KID} $\downarrow$ & \textbf{CLIP-S} $\uparrow$  & \textbf{PD} $\downarrow$ & \textbf{SD} $\downarrow$ \\
\midrule
 & & \cmark & & & & 12.73 & 3.59 & 30.24 & 8.64 & 0.643 \\
 & & \cmark & \cmark & & & \bf12.40 & \bf3.36 & 30.34 & 7.53 & 0.435 \\
\rowcolor{blond}
\textbf{\ours (ours)} & & \cmark & \cmark & \cmark & & 12.81 & 3.86 & \bf30.75 & \bf7.22 & \bf0.317 \\
\bottomrule
\end{tabular}
}
\end{center}
\vspace{-0.4cm}
\caption{Performance analysis on the unpaired setting of both datasets as input modalities vary.}
\label{tab:modalities_unpaired}
\vspace{-.35cm}
\end{table}

\begin{table*}[t]
\begin{center}
\footnotesize
\setlength{\tabcolsep}{.35em}
\begin{tabular}{llc c ccc c ccccc c ccccc}
\toprule
& & & & \multicolumn{3}{c}{\textbf{Modalities}} & & \multicolumn{5}{c}{\textbf{\dataset}} & \multicolumn{5}{c}{\textbf{\datasetviton}} \\
\cmidrule{5-7} \cmidrule{9-13} \cmidrule{15-19} 
\multicolumn{2}{l}{\textbf{Model}} & \textbf{Resolution} & & \textbf{Text} & \textbf{Pose} & \textbf{Sketch} & & 
\textbf{FID} $\downarrow$ & \textbf{KID} $\downarrow$ & \textbf{CLIP-S} $\uparrow$  & \textbf{PD} $\downarrow$ & \textbf{SD} $\downarrow$ &  &
\textbf{FID} $\downarrow$ & \textbf{KID} $\downarrow$ & \textbf{CLIP-S} $\uparrow$  & \textbf{PD} $\downarrow$ & \textbf{SD} $\downarrow$ \\
\midrule
\multicolumn{2}{l}{\textit{Paired setting}} \\
& ControlNet~\cite{zhang2023adding} & 512$\times$384 & & \cmark & \cmark & & & 18.36 & 9.82 & 29.00 & 7.46 & 0.462 & & 19.08 & 9.35 & 30.03 & 7.72 & 0.392\\
\rowcolor{blond}
& \textbf{\ours (ours)} & 512$\times$384 & & \cmark & \cmark & & & \textbf{6.31} & \textbf{2.33} & \textbf{31.67} & \textbf{5.31} & \textbf{0.405} & & \textbf{11.07} & \textbf{3.36} & \textbf{32.27} & \textbf{6.77} & \textbf{0.318}\\
\cmidrule{2-18}
& ControlNet~\cite{zhang2023adding} & 512$\times$384 & & \cmark &  & \cmark & & 27.23 & 19.01 & 27.07 & 7.54 & 0.436 & & 25.44 & 17.05 & 28.31 & 8.16 & 0.298\\
\rowcolor{blond}
& \textbf{\ours (ours)} & 512$\times$384 & & \cmark & & \cmark & & \textbf{5.72} & \textbf{2.15} & \textbf{31.69} & \textbf{4.94} & \textbf{0.373} & & \textbf{10.64} & \textbf{3.26} & \textbf{32.31} & \textbf{6.18} & \textbf{0.255} \\
\midrule
\multicolumn{2}{l}{\textit{Unpaired setting}} \\
& ControlNet~\cite{zhang2023adding} & 512$\times$384 & & \cmark & \cmark & & & 20.66 & 11.58 & 27.57 & 8.15 & 0.577 & & 21.03 & 10.34 & 28.11 & 8.38 & 0.534\\
\rowcolor{blond}
& \textbf{\ours (ours)} & 512$\times$384 & & \cmark & \cmark & & & \textbf{7.82} & \textbf{2.85} & \textbf{29.93} & \textbf{6.26} & \textbf{0.519} & & \textbf{12.40} & \textbf{3.36} & \textbf{30.34} & \textbf{7.53} & \textbf{0.435}\\
\cmidrule{2-18}
& ControlNet~\cite{zhang2023adding} & 512$\times$384 & & \cmark &  & \cmark & & 29.61 & 20.83 & 25.75 & 9.74 & 0.544 & & 27.41 & 18.66 & 26.63 & 9.53 & 0.416 \\
\rowcolor{blond}
& \textbf{\ours (ours)} & 512$\times$384 & & \cmark & & \cmark & & \textbf{7.65} & \textbf{2.70} & \textbf{30.21} & \textbf{7.50} & \textbf{0.456} & & \textbf{12.65} & \textbf{3.59} & \textbf{30.69} & \textbf{7.49} & \textbf{0.320} \\
\bottomrule
\end{tabular}
\end{center}
\vspace{-0.4cm}
\caption{Performance comparison with ControlNet on the \dataset and \datasetviton datasets for both paired and unpaired settings.
}
\vspace{-0.3cm}
\label{tab:controlnet}
\end{table*}

In Table~\ref{tab:modalities_unpaired}, we show the performance of our \ours model when masking different input modalities. In this case, we report the results on the unpaired setting of both datasets. As it can be seen, evaluation metrics measuring the realism of the generation (\ie~FID and KID) are comparable among different cases, while the pose distance and sketch distance metrics correlate in general with the given input (\ie~with the pose map and the garment sketch, respectively). Moreover, in this case, the warped in-shop garment not fitting the model's body shape affects the pose distance metric for the \dataset dataset.

Finally, in Table~\ref{tab:controlnet} we report a comparison with the concurrent work ControlNet~\cite{zhang2023adding} adapted to work with the Stable Diffusion inpaint denoising network. Following the original paper, we only condition ControlNet on text plus an additional modality (\ie~pose or sketch). It is worth noting that across all configurations, \ours outperforms ControlNet by a significant margin.

\begin{figure}[!t]
\resizebox{\linewidth}{!}{
\centering
\includegraphics[width=\linewidth]{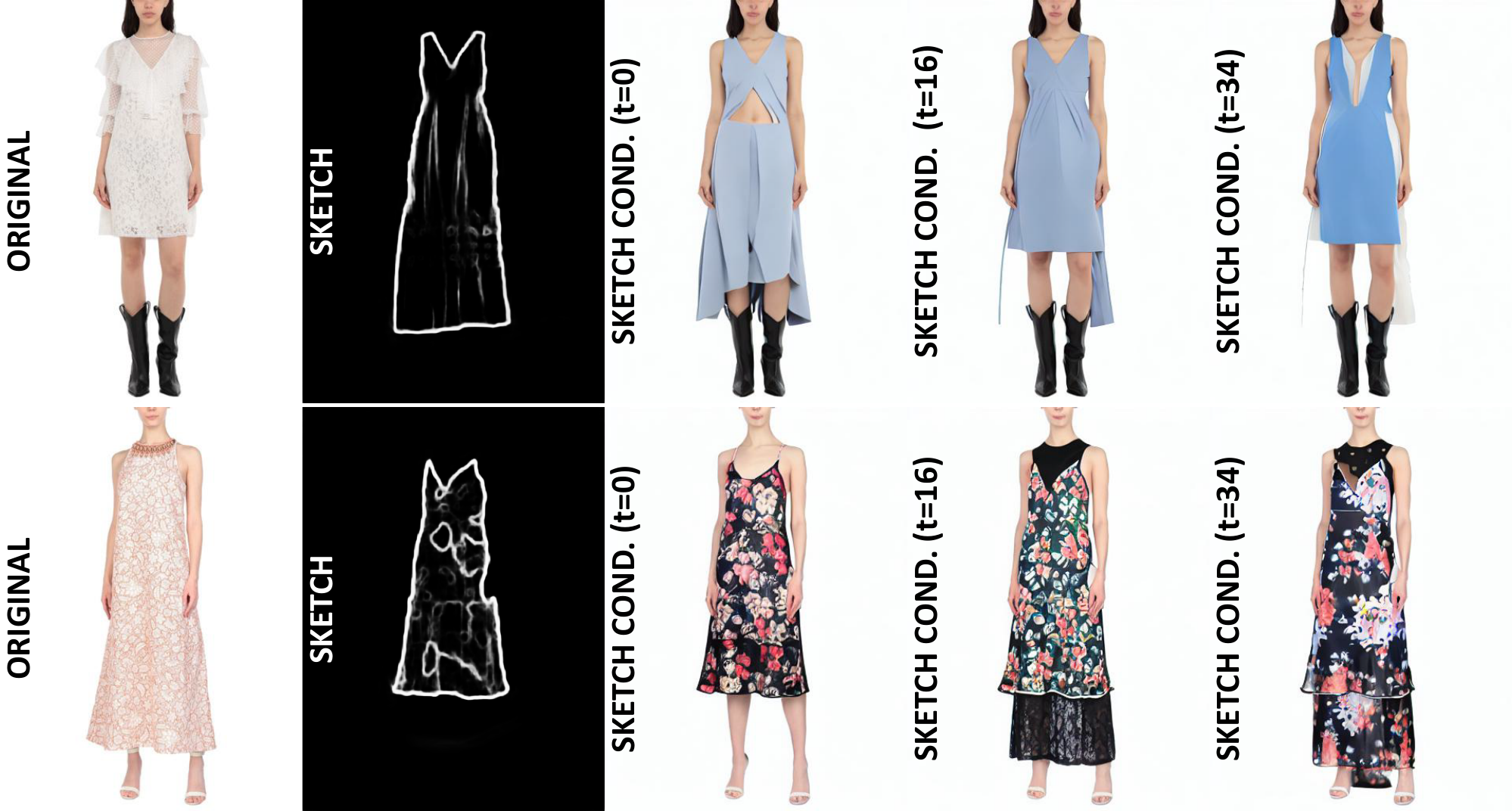}
}
\vspace{-0.4cm}
\caption{Time window conditioned examples on \dataset. We report qualitative results fixing the sketch conditioning steps to around a third of diffusion steps and varying the starting conditioning timestep (\ie~$t_{start}=0,16,34$).}
\label{fig:dresscode_timewindow}
\vspace{-0.3cm}
\end{figure}

\begin{figure*}
\begin{center}
\resizebox{0.98\linewidth}{!}{
\includegraphics[width=\linewidth]{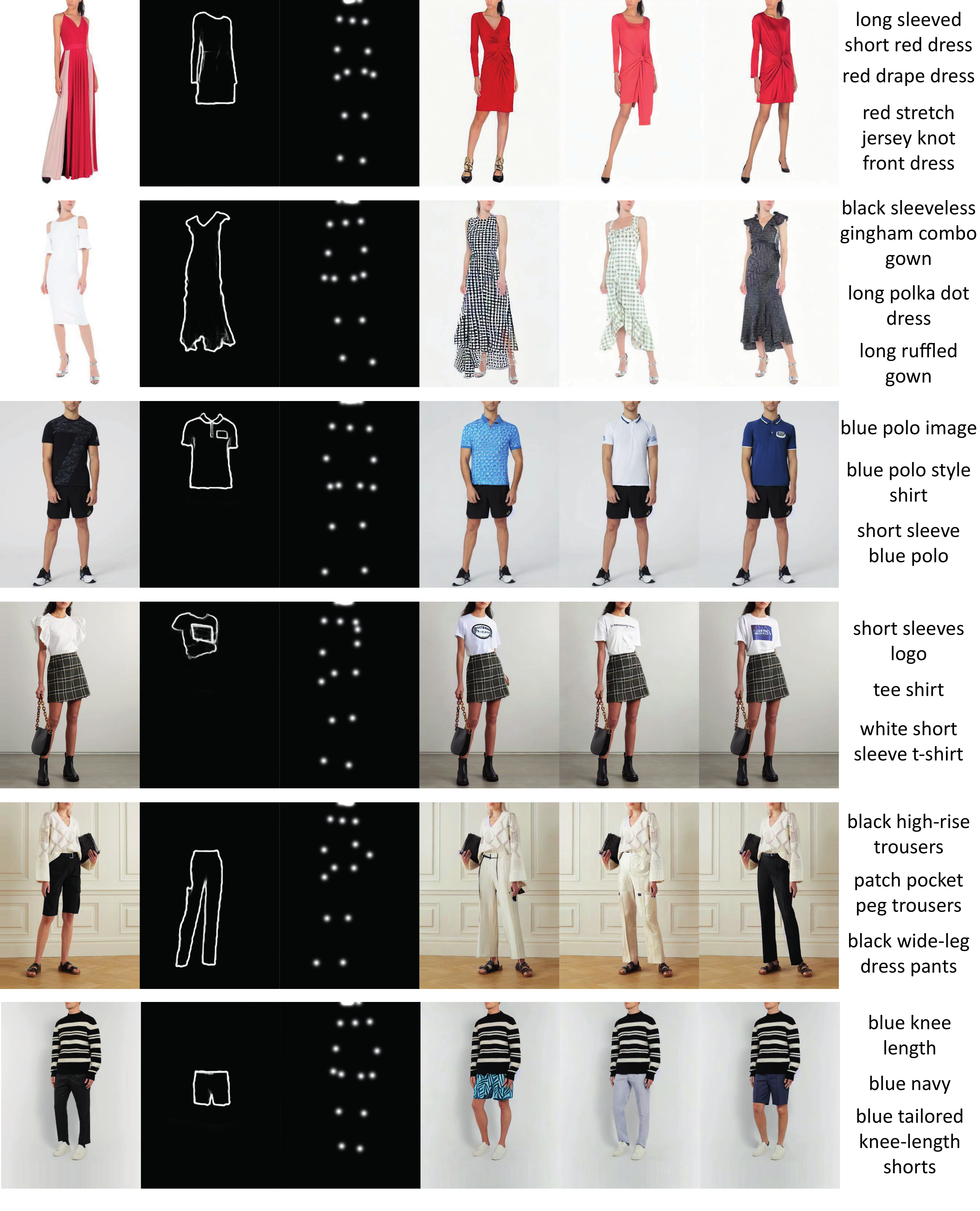}
}
\end{center}
\vspace{-0.6cm}
\caption{Qualitative comparison on \dataset. From left to right: model's image, input sketch, pose map, image generated by Stable Diffusion~\cite{rombach2022high}, image generated by SDedit~\cite{meng2022sdedit}, image generated by MGD (ours), and noun chunks.}
\label{fig:dresscode_competitors}
\end{figure*}

\begin{figure*}
\begin{center}
\resizebox{0.98\linewidth}{!}{
    \includegraphics[width=\linewidth]{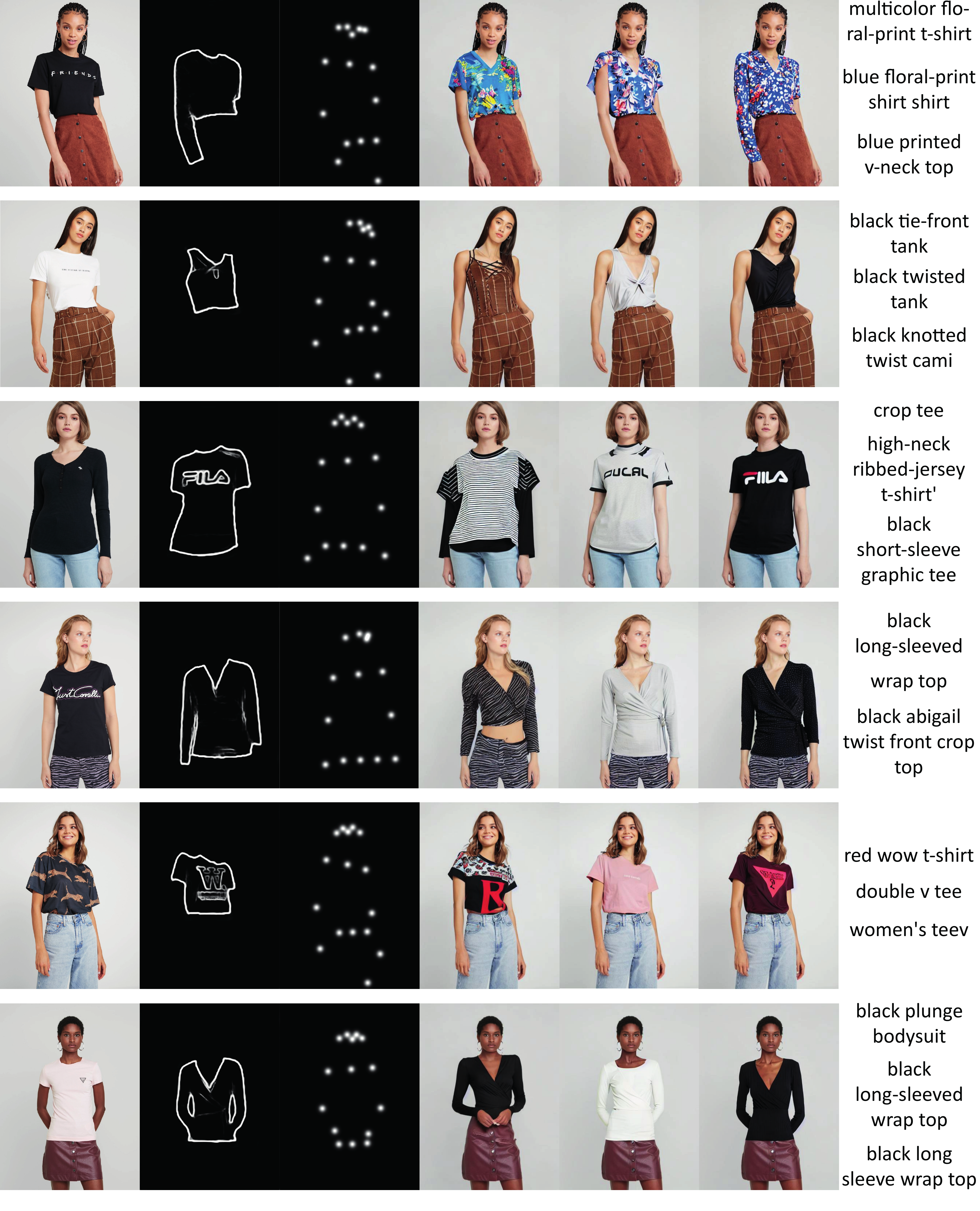}
}
\end{center}
\vspace{-0.6cm}
\caption{Qualitative comparison on \datasetviton. From left to right: model's image, input sketch, pose map, image generated by Stable Diffusion~\cite{rombach2022high}, image generated by SDedit~\cite{meng2022sdedit}, image generated by MGD (ours), and noun chunks.}
\label{fig:vitonhd_competitors}
\end{figure*}

\begin{figure*}
\begin{center}
\resizebox{0.98\linewidth}{!}{
\includegraphics[width=\linewidth]{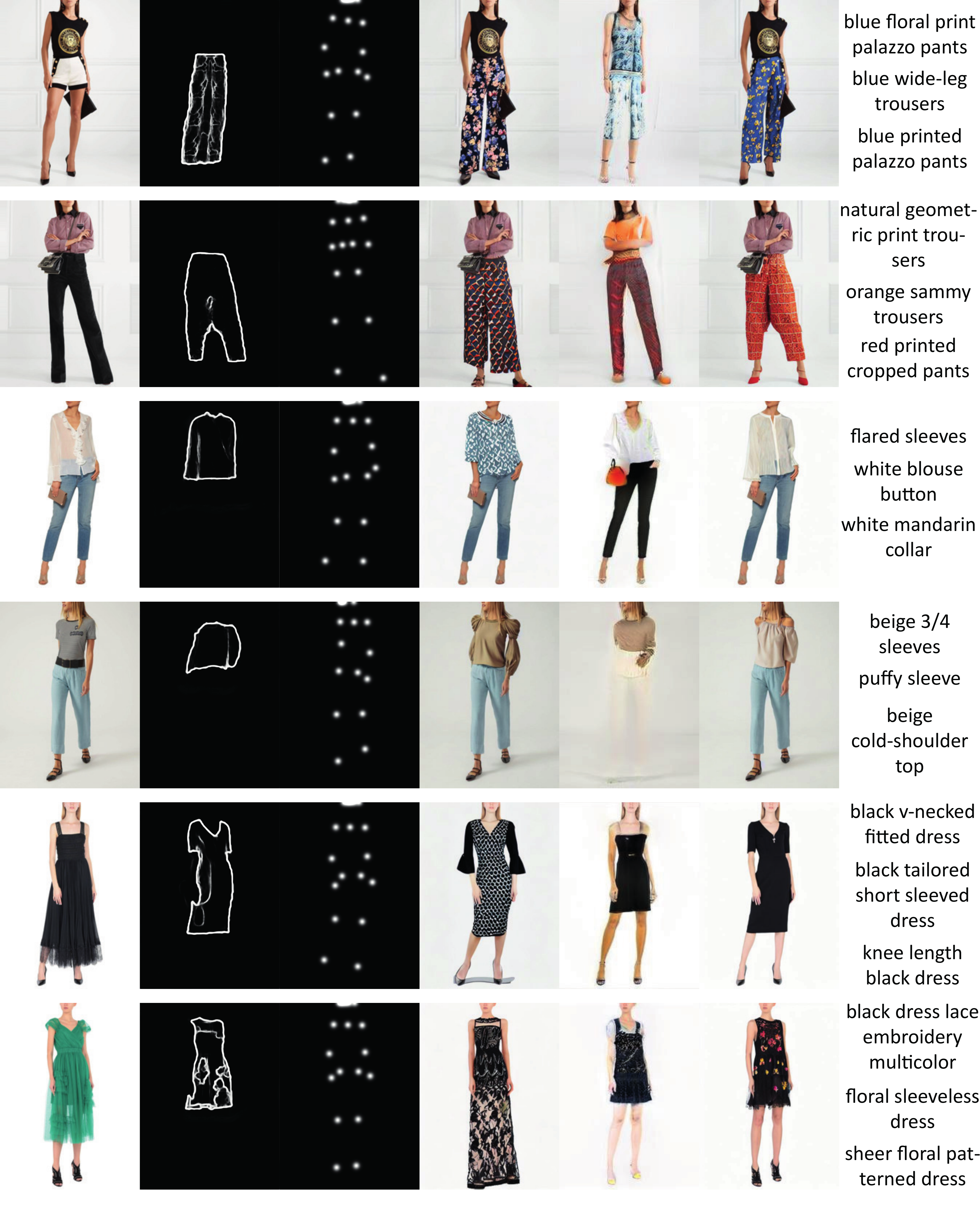}
}
\end{center}
\vspace{-0.6cm}
\caption{Qualitative comparison with low-resolution images on \dataset. From left to right: model's image, input sketch, pose map, image generated by Stable Diffusion~\cite{rombach2022high}, image generated by FICE~\cite{pernuvs2023fice}, image generated by MGD (ours), and noun chunks.}
\label{fig:dresscode_lowres}
\end{figure*}

\begin{figure*}
\begin{center}
\resizebox{0.98\linewidth}{!}{
    \includegraphics[width=\linewidth]{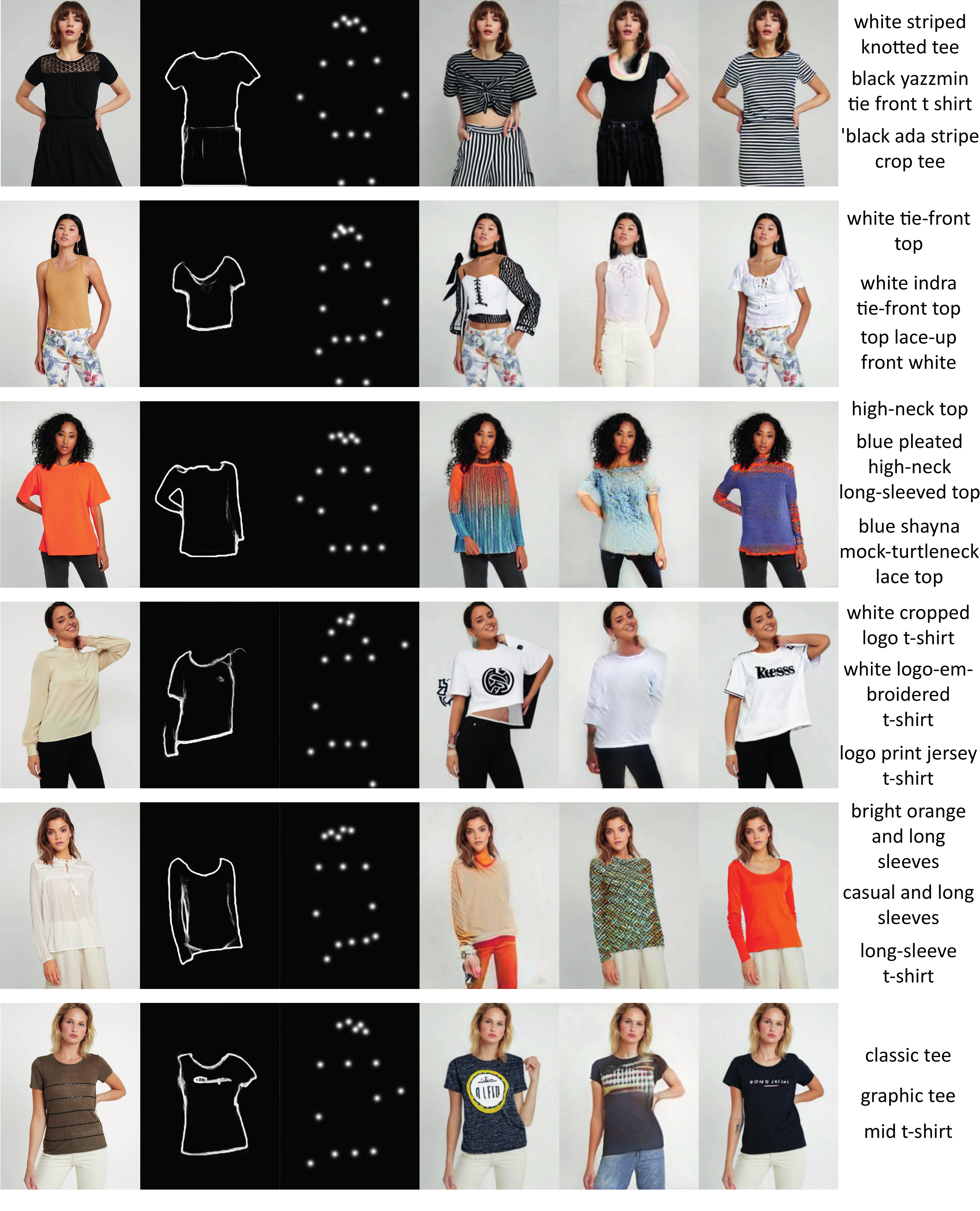}
}
\end{center}
\vspace{-0.6cm}
\caption{Qualitative comparison with low-resolution images on \datasetviton. From left to right: model's image, input sketch, pose map, image generated by Stable Diffusion~\cite{rombach2022high}, image generated by FICE~\cite{pernuvs2023fice}, image generated by MGD (ours), and noun chunks.}
\label{fig:vitonhd_lowres}
\end{figure*}

\begin{figure*}
\begin{center}
\resizebox{\linewidth}{!}{
    \includegraphics[width=\linewidth]{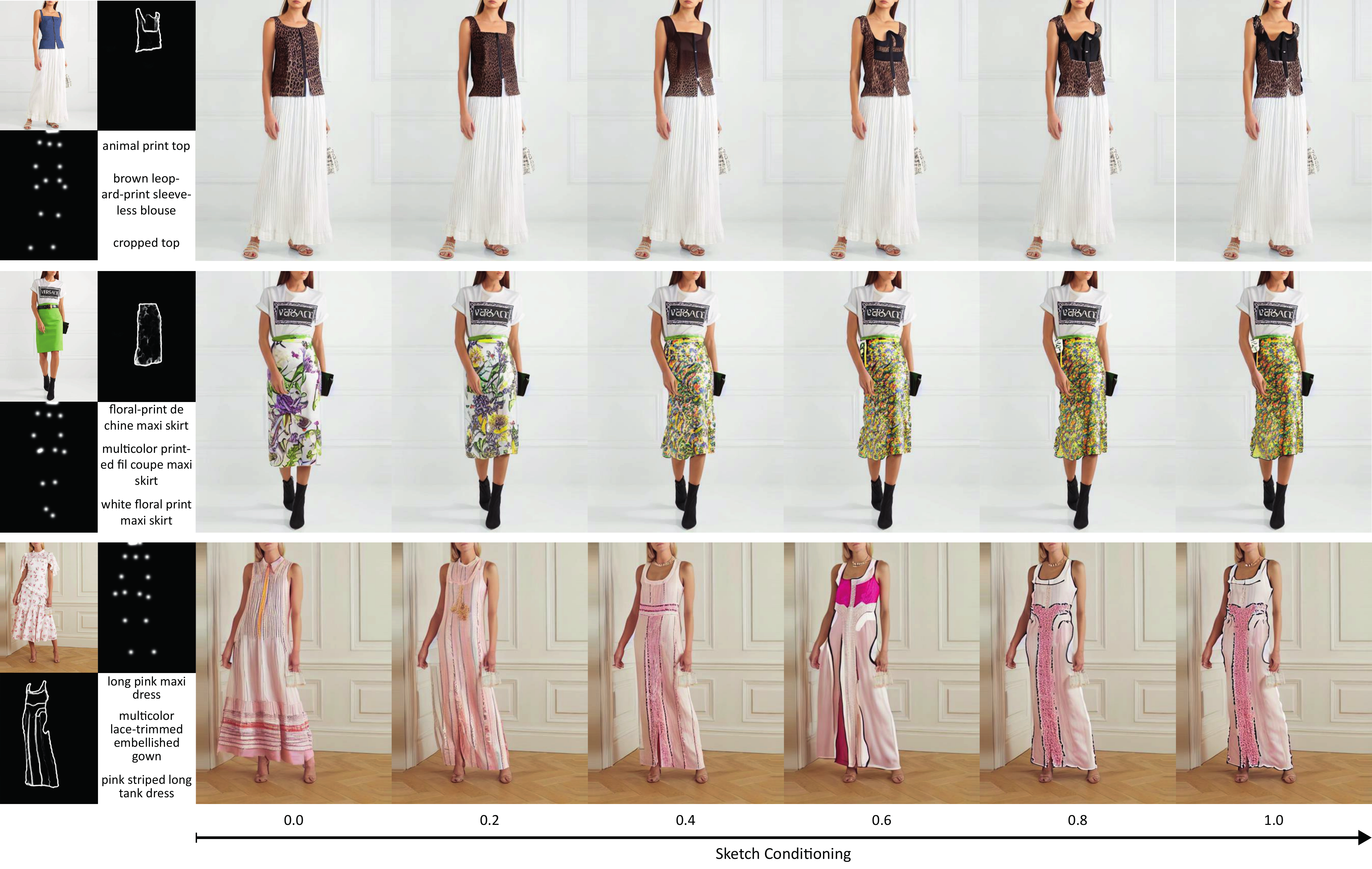}
}
\end{center}
\vspace{-0.6cm}
\caption{Qualitative results generated by \ours increasing the sketch conditioning steps.}
\label{fig:sketchcond_ablation}
\vspace{0.5cm}
\end{figure*}

\tit{Qualitative results}
We also show additional qualitative results for both datasets. Specifically, in Fig.~\ref{fig:dresscode_competitors} and Fig.~\ref{fig:vitonhd_competitors}, we compare images generated by our approach and competitors using a resolution of $512\times384$, for \dataset and \datasetviton, respectively. Instead, in Fig.~\ref{fig:dresscode_lowres} and Fig.~\ref{fig:vitonhd_lowres}, we report low-resolution qualitative comparisons. 
Fig.~\ref{fig:sketchcond_ablation} shows some qualitative results varying the sketch conditioning parameter. Increasing the number of sketch conditioning steps leads to images that better follow the given sketch while slightly reducing the realism of the generated garments.
Finally, we investigate the conditioning contribution in various time windows in Fig.~\ref{fig:dresscode_timewindow}. We perform this experiment by fixing the sketch conditioning steps to around a third of diffusion steps and varying the starting conditioning timestep (\ie~$t_{start}=0,16,34$). Qualitative results show that starting the sketch conditioning in the central (\ie~$t_{start}=16$, $t_{end}=34$) or final denoising steps (\ie~$t_{start}=34$, $t_{end}=50$) leads the model to generate images that do not follow the input sketch and present artifacts.

\begin{figure*}
\begin{center}
\resizebox{0.98\linewidth}{!}{
    \includegraphics[width=\linewidth]{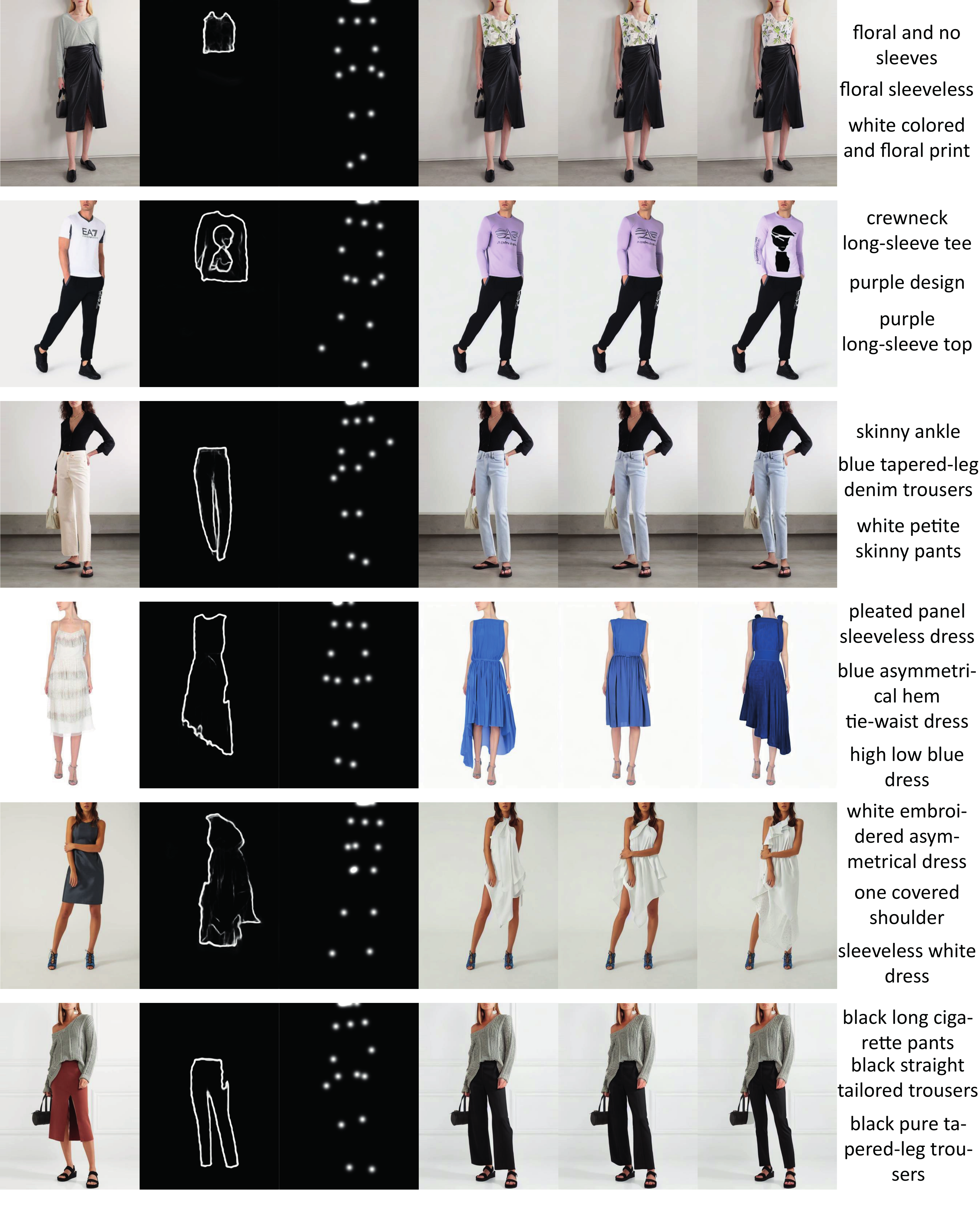}
}
\end{center}
\vspace{-0.6cm}
\caption{Qualitative comparison of images generated by our model on \dataset using different conditioning modalities. From left to right: model's image, input sketch, pose map, image generated using only text, image generated using text and pose map, image generated with all input modalities (\ie~text, pose map, and sketch).}
\label{fig:dresscode_modalities}
\end{figure*}

\begin{figure*}
\begin{center}
\resizebox{\linewidth}{!}{
    \includegraphics[width=\linewidth]{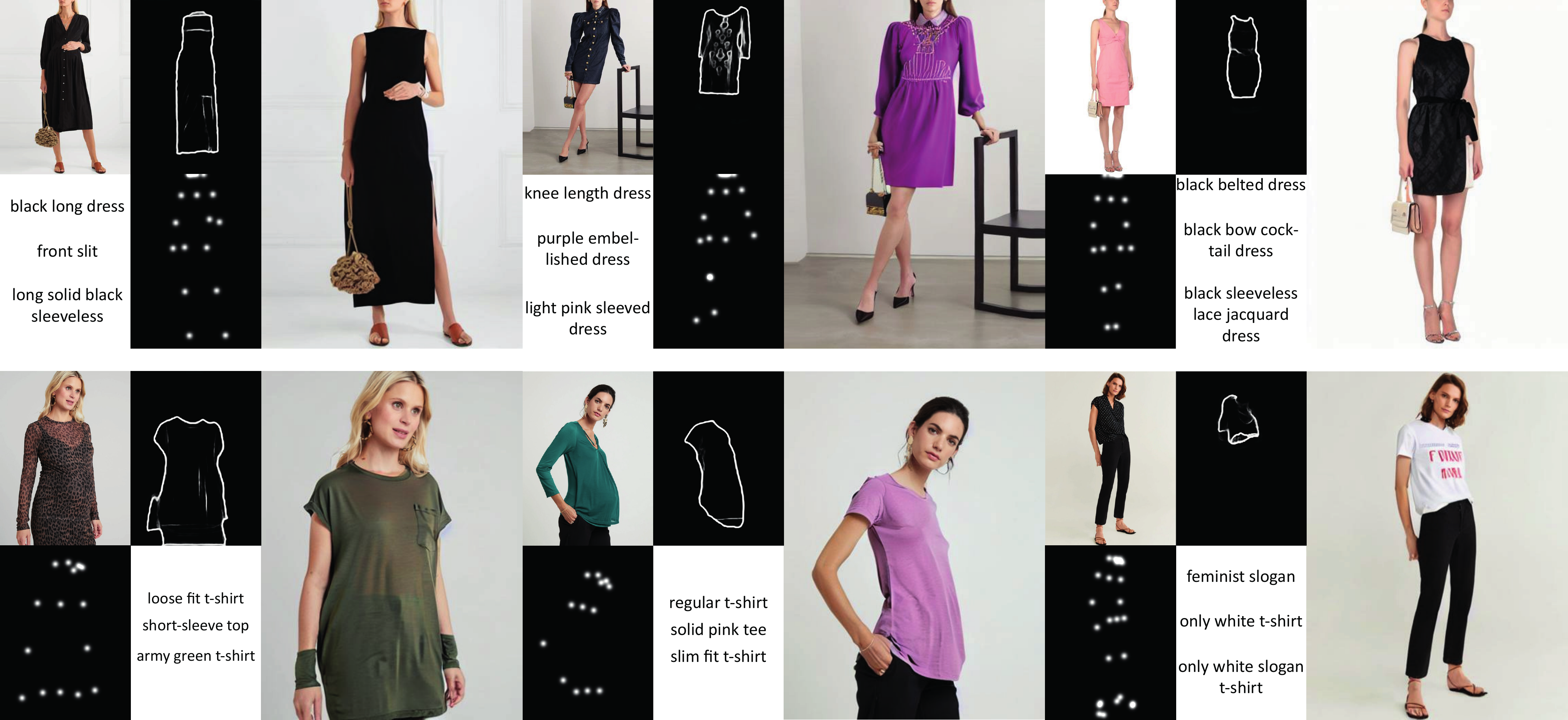}
}
\end{center}
\vspace{-0.3cm}
\caption{Failure cases on \dataset (first row) and \datasetviton (second row).}
\label{fig:failure_cases}
\vspace{0.5cm}
\end{figure*}

\tit{Limitations and failure cases}
Fig.~\ref{fig:failure_cases} shows some failure cases of the proposed approach. In the first row, the first two examples show that our model sometimes fails to generate hands accurately when they occupy a limited area within the source image. This behavior is intrinsic in LDMs family~\cite{rombach2022high} and derives from the high spatial compression nature of the latent space ($8 \times$ for each spatial dimension). Instead, the third example of the first row and the first two samples of the second row highlight the dependence of our model performance from the given sketch. When the geometric warping module fails to generate a sketch able to fit the model's shape, the generation task fails as well, creating unwanted artifacts (\eg~a sketch may be smaller than the model's body shape as in the third example of the first row, resulting in an artifact near the model's left hand).

\end{document}